\documentclass[final]{cvpr}
\usepackage{times}
\usepackage{epsfig}
\usepackage{graphicx}
\usepackage{amsmath}
\usepackage{amssymb}
\usepackage{color}
\usepackage{multirow}
\usepackage{diagbox}
\usepackage{array}
\usepackage{booktabs}
\usepackage{overpic}
\usepackage[table]{xcolor}

\usepackage{eucal}

\newcommand{\tabincell}[2]{\begin{tabular}{@{}#1@{}}#2\end{tabular}}  
\usepackage{pifont}
\usepackage{overpic}
\newcommand{\cmark}{\ding{51}}%
\newcommand{\xmark}{\ding{55}}%

\usepackage{algorithm}
\usepackage{algorithmicx}
\usepackage{algpseudocode}
\usepackage{booktabs}       
\makeatletter
\def\wideubar{\underaccent{{\cc@style\underline{\mskip10mu}}}}
\def\Wideubar{\underaccent{{\cc@style\underline{\mskip8mu}}}}
\makeatother
\usepackage{subfig}
\newcommand{\tablestyle}[2]{\setlength{\tabcolsep}{#1}\renewcommand{\arraystretch}{#2}\centering\footnotesize}
\newcommand{\egno}{\textit{e}.\textit{g}.} 
\newcommand{\etcno}{\textit{etc}} 
\newcommand{\ieno}{\textit{i}.\textit{e}.}

\makeatletter
\def\widebar{\accentset{{\cc@style\underline{\mskip10mu}}}}
\def\Widebar{\accentset{{\cc@style\underline{\mskip8mu}}}}
\makeatother

\usepackage{enumitem}
\usepackage{caption}
\usepackage{comment}
\usepackage{xcolor}

\begin{document}

\title{Style Normalization and Restitution for Generalizable Person Re-identification}

\author{{Xin Jin{$^{1}$}\thanks{This work was done when Xin Jin was an intern at Microsoft Research Asia.}} \qquad Cuiling Lan{$^{2}$}\thanks{Corresponding Author.} \qquad   Wenjun Zeng{$^{2}$} \qquad  Zhibo Chen{$^{1\dag}$} \qquad Li Zhang{$^{3}$}\\
	\normalsize
	$^{1}$\	University of Science and Technology of China ~~ $^{2}$\,Microsoft Research Asia, Beijing, China ~~ $^{3}$\ University of Oxford\\
	\normalsize
	{\tt\small jinxustc@mail.ustc.edu.cn\quad \{culan,wezeng\}@microsoft.com\quad chenzhibo@ustc.edu.cn\quad
	lz@robots.ox.ac.uk}
	}
	

\maketitle
\thispagestyle{empty} 

\begin{abstract}
  Existing fully-supervised person re-identification (ReID) methods usually 
  suffer from poor generalization capability caused by domain gaps. 
  The key to solving this problem lies in filtering out \textbf{identity-irrelevant} interference and learning domain-invariant person representations.
  In this paper, we aim to design a generalizable person ReID framework which trains a model on source domains yet is able to generalize/perform well on target domains. To achieve this goal, we propose a simple yet effective \textbf{S}tyle \textbf{N}ormalization and \textbf{R}estitution (SNR) module. Specifically, we filter out style variations (\eg, illumination, color contrast) by Instance Normalization (IN). However, such a process inevitably removes discriminative information. We propose to distill identity-relevant feature from the removed information and restitute it to the network to ensure high discrimination. For better disentanglement, we enforce a dual causality loss constraint in SNR to encourage the separation of identity-relevant features and identity-irrelevant features. Extensive experiments demonstrate the strong generalization capability of our framework. Our models empowered by the SNR modules significantly outperform the state-of-the-art domain generalization approaches on multiple widely-used person ReID benchmarks, and also show superiority on unsupervised domain adaptation.
\end{abstract}
\vspace{-4mm}

\section{Introduction}

\begin{figure}
  \centerline{\includegraphics[width=1.0\linewidth]{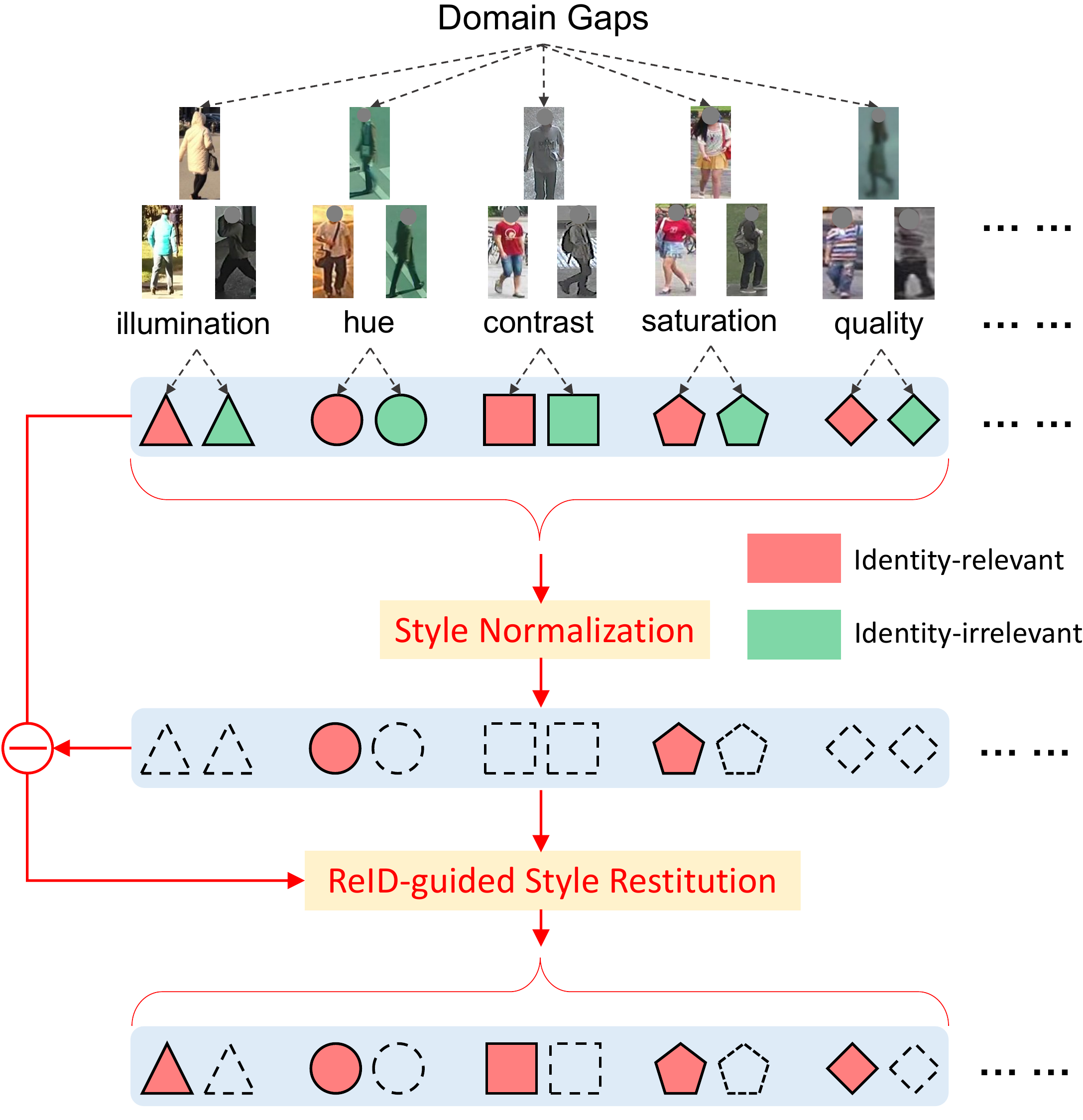}}
  \vspace{-2mm}
  \caption{Illustration of motivation and our idea. Person images captured from different cameras and environments present style variations which result in domain gaps. We use style normalization (with Instance Normalization) to alleviate style variations. However, this also results in the loss of some discriminative (identity-relevant) information. We propose to further restitute such information from the residual of the original information and the normalized information for generalizable and discriminative person ReID.}
\label{fig:motivation}
\vspace{-6mm}
\end{figure}
Person re-identification (ReID) aims at matching/identifying a specific person across cameras, times, and locations. It facilitates many applications and has attracted a lot of attention. 

Abundant approaches have been proposed for supervised person ReID, where a model is trained and tested on different splits of the same dataset \cite{zhang2016learning,su2017pose,zhao2017spindle,ge2018fd,qian2018pose,zhang2019DSA,jin2020semantics,jin2020uncertainty}. They typically focus on addressing the challenge of geometric misalignment among images caused by diversity of poses/viewpoints. In general, they perform well on the trained dataset but suffer from signiﬁcant performance degradation (poor generalization capability) when testing on a previously unseen dataset. There are usually style discrepancies across domains/datasets which hinder the achievement of high generalization capability. Figure \ref{fig:motivation} shows some example images\footnote{All faces in the images are masked for anonymization.} from different ReID datasets. The person images are captured by different cameras under different environments (\egno, lighting, seasons). They present a large style discrepancy in terms of illumination, hue, color contrast and saturation, quality/resolution, \etcno. For a ReID system, we expect it to be able to identify the same person even captured in different environments, and distinguish between different people even if their appearance are similar. Both generalization and discrimination capabilities, although seemly conflicting with each other, are very important for robust ReID.

Considering the existence of domain gaps and poor generalization capability, fully-supervised approaches or settings are not practical for real-world widespread ReID system deployment, where the onsite manual annotation on the target domain data is expensive and hardly feasible. In recent years, some unsupervised domain adaptation (UDA) methods have been studied to adapt a ReID model from source to target domain \cite{wang2018transferable,tang2019unsupervised,liu2019adaptive,qi2019novel,fan2018unsupervised,yu2019unsupervised,yang2019patch}. UDA models update using \emph{unlabeled} target domain data, emancipating the labelling efforts. However, data collection and model update are still required, adding additional cost. 

We mainly focus on the more economical and practical domain generalizable person ReID. Domain generalization (DG) aims to design models that are generalizable to previously unseen domains \cite{muandet2013domain,jia2019frustratingly,song2019generalizable}, without having to access the target domain data and labels, and without requiring model updating. Most DG methods assume that the source and target domains have the same label space \cite{khosla2012undoing,li2018learning,muandet2013domain,shankar2018generalizing} and they are not applicable to ReID since the target domains for ReID typically have a different label space from the source domains.
Generalizable person ReID is challenging which aims to achieve high discrimination capability on \emph{unseen} target domain that may have large domain discrepancy. The study on domain generalizable ReID is rare \cite{song2019generalizable,jia2019frustratingly} and remains an open problem. Jia \etal \cite{jia2019frustratingly} and Zhou \etal \cite{zhou2019omni} integrate Instance Normalization (IN) in the networks to alleviate the domain discrepancy due to appearance style variations. However, IN inevitably results in the loss of some discriminative features \cite{huang2017arbitrary,pan2018two}, hindering the achievement of high efficiency ReID.





\begin{figure*}
  \centerline{\includegraphics[width=1.0\linewidth]{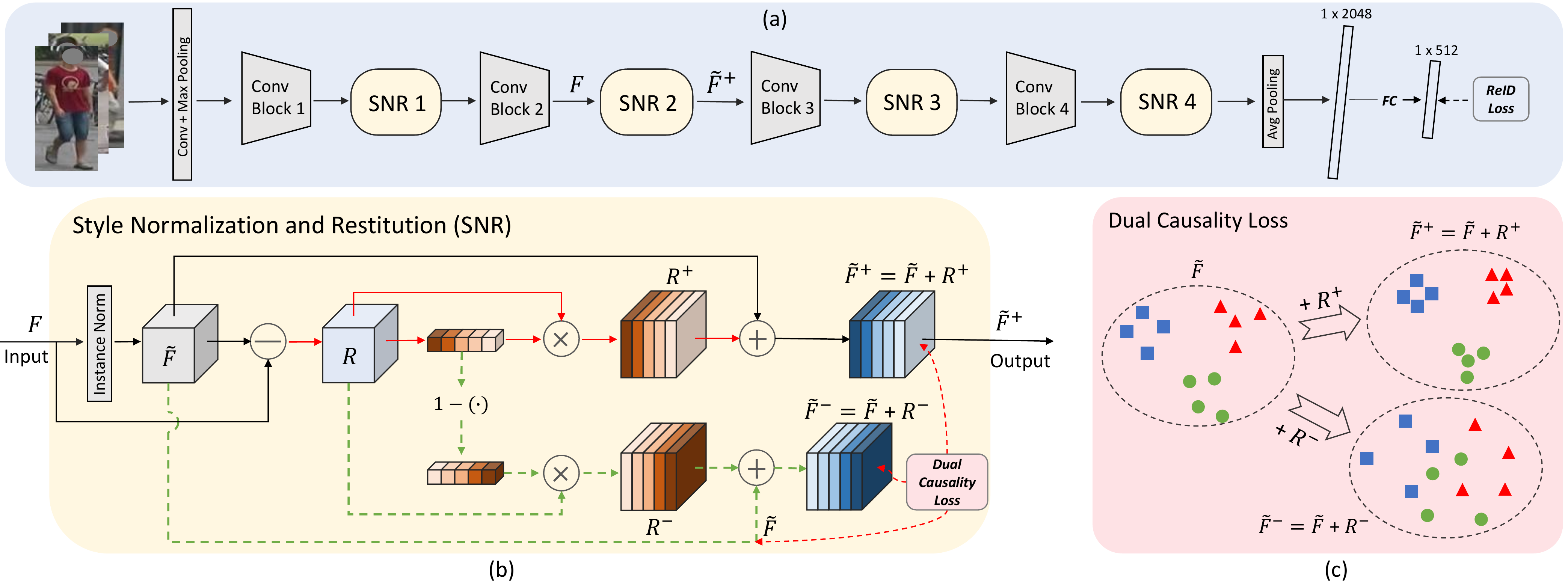}}
  \vspace{-2mm}
  \caption{Overall flowchart. (a) Our generalizable person ReID network with the proposed Style Normalization and Restitution (SNR) module being plugged in after some convolutional blocks. Here, we use ResNet-50 as our backbone for illustration. (b) Proposed SNR module. Instance Normalization (IN) is used to eliminate some style discrepancies followed by identity-relevant feature restitution (marked by red solid arrows). Note the branch with dashed green line is only used for enforcing loss constraint and is discarded in inference. (c) Dual causality loss constraint encourages the disentanglement of a residual feature $R$ to identity-relevant one ($R^+$) and identity-irrelevant one ($R^-$), which enhances and decreases, respectively, the discrimination by adding them to the style normalized feature $\widetilde{F}$.}
\label{fig:flowchart}
\vspace{-3.5mm}
\end{figure*}

In this paper, we aim to design a generalizable ReID framework which achieves both high generalization capability and discrimination capability. The key is to find a way to disentangle the identity-relevant features and the identity-irrelevant features (\eg, image styles). Figure \ref{fig:motivation} illustrates our main idea. Considering the domain gaps among image samples, we perform style normalization by means of IN to eliminate style variations. However, the normalization inevitably discards some discriminative information and thus may hamper the ReID performance. From the residual information (which is the difference between the original information and the normalized information), we further distill the identity-relevant information as a compensation to the normalized information. Figure \ref{fig:flowchart} shows our framework with the proposed Style Normalization and Restitution (SNR) modules embedded. To better disentangle the identity-relevant features from the residual, a dual causality loss constraint is added by ensuring the features after restitution of identity-relevant features to be more discriminative, and the features after compensation of identity-irrelevant features to be less discriminative.  




We summarize our main contributions as follows:
\begin{itemize}[leftmargin=*,noitemsep,nolistsep]

\item We propose a practical domain generalizable person ReID framework that generalizes well on previously unseen domains/datasets. Particularly, we design a Style Normalization and Restitution (SNR) module. SNR is simple yet effective and can be used as a \emph{plug-and-play} module for existing ReID architectures to enhance their generalization capabilities. 

\item To facilitate the restitution of identity-relevant features from those discarded in the style normalization phase, we introduce a dual causality loss constraint in SNR for better feature disentanglement.
\end{itemize}

We validate the effectiveness of the proposed SNR module on multiple widely-used benchmarks and settings. Our models significantly outperform the state-of-the-art domain generalizable person ReID approaches and can also boost the performance of unsupervised domain adaptation for ReID. 

\section{Related Work}

\noindent\textbf{Supervised Person ReID.} In the last decade, fully-supervised person ReID has achieved great progress, especially for deep learning based approaches \cite{su2017pose,li2017learning,zhao2017spindle,ge2018fd,qian2018pose,zhang2019DSA}. These methods usually perform well on the testing set of the source  datasets but generalize poorly to previously unseen domains/datasets due to the style discrepancy across domains. This is problematic especially in practical applications, where the target scenes typically have different styles from the source domains and there is no readily available target domain data or annotation for training.  




\noindent\textbf{Unsupervised Domain Adaptation (UDA) for Person ReID.} When the target domain data is accessible, even without annotations, it can be explored for the domain adaptation for enhancing the ReID performance. This requires target domain data collection and model updating. UDA-based ReID methods can be roughly divided into three categories: \emph{style transfer} \cite{deng2018image,wei2018person,liu2019adaptive}, \emph{attribute recognition} \cite{wang2018transferable,yu2017cross,qi2019novel}, and \emph{target-domain pseudo label estimation} \cite{fan2018unsupervised,song2018unsupervised,zhong2018generalizing,tang2019unsupervised,zhang2019self,yu2019unsupervised}. For pseudo label estimation, recently, Yu \etal propose a method called \textbf{m}ultil\textbf{a}bel \textbf{r}eference learning (MAR) which evaluates the similarity of a pair of images by comparing them to a set of known reference persons to mine hard negative samples \cite{yu2019unsupervised}.

Our proposed domain generalizable SNR module can also be combined with the UDA methods (\egno, by plugging into the UDA backbone) to further enhance the ReID performance. We will demonstrate its effectiveness by combining it with the UDA approach of MAR in Subsection \ref{subsec:SOTA}.

\noindent\textbf{Domain Generalization (DG).}
Domain Generalization is a challenging problem of learning models that is generalizable to unseen domains \cite{muandet2013domain,shankar2018generalizing}. Muandet \etal learn an invariant transformation by minimizing the dissimilarity across source domains \cite{muandet2013domain}. A learning-theoretic analysis shows that reducing dissimilarity improves the generalization ability on new domains. CrossGrad \cite{shankar2018generalizing} generates pseudo training instances by pertubations in the loss gradients of the domain classifier and category classifier respectively. Most DG methods assume that the source and target domains have the same label space. However, ReID is an open-set problem where the target domains typically have different identities from the source domains, so that the general DG methods could not be directly applied to ReID.

Recently, a strong baseline for domain generalizable person ReID is proposed by simply combing multiple source datasets and training a single CNN \cite{kumar2019fairest}. Song \etal \cite{song2019generalizable} propose a generalizable person ReID framework by using a meta-learning pipeline to make the model domain invariant. To overcome the inconsistency of label spaces among different datasets, it maintains a training datasets shared memory bank. Instance Normalization (IN) has been widely used in image style transfer \cite{huang2017arbitrary,ulyanov2017improved} and proved that it actually performs a kind of style normalization \cite{pan2018two,huang2017arbitrary}. Jia \etal \cite{jia2019frustratingly} and Zhou \etal \cite{zhou2019omni} apply this idea to ReID to alleviate the domain discrepancy and boost the generalization capability. However, IN inevitably discards some discriminative information. In this paper, we study how to design a generalibale ReID framework that can exploit the merit of IN while avoiding the loss of discriminative information.

\vspace{-1mm}
\section{Proposed Generalizable Person ReID}
\vspace{-1mm}
We aim at designing a generalizable and robust person ReID framework. During the training, we have access to one or several annotated source datasets. The trained model will be deployed directly to unseen domains/datasets and is expected to work well with high generalization capability. 

Figure \ref{fig:flowchart} shows the overall flowchat of our framework. Particularly, we propose a Style Normalization and Restitution (SNR) module to boost the generalization and discrimination capability of ReID models especially on unseen domains. SNR can be used as a plug-and-play module for existing ReID networks. Taking the widely used ReID network of ResNet-50 \cite{he2016deep,almazan2018re,luo2019bag} as an example (see Figure \ref{fig:flowchart}(a)), SNR module is  added after each convolutional block. In the SNR module, we first eliminate style discrepancy among samples by Instance Normalization (IN). Then, a dedicated restitution step is proposed to distill identity-relevant (discriminative) features from those previsouly discarded by IN, and add them to the normalized features. Moreover, for the SNR module, we design a dual causality loss constraint to facilitate the distillation of identity-relevant features from the information discarded by IN.

\subsection{Style Normalization and Restitution (SNR)}
\label{subsec:SNR}

Person images for ReID could be captured by different cameras under different scenes and environments (\egno, indoor/outdoors, shopping malls, street, sunny/cloudy). As shown in Figure \ref{fig:motivation}, they present style discrepancies (\egno, in illumination, hue, contrast, saturation, quality), especially for samples from two different datasets/domains. Domain discrepancy between the source and target domain generally hinders the generalization capability of ReID models. 

A learning-theoretic analysis shows that reducing dissimilarity improves the generalization ability on new domains \cite{muandet2013domain}. Instance Normalization (IN) performs some kinds of style normalization which reduces the discrepancy/dissimilarity among instances/samples \cite{huang2017arbitrary,pan2018two}, so it can enhance the generalization ability of networks \cite{pan2018two,jia2019frustratingly,zhou2019omni}. However, IN inevitably removes some discriminative information and results in weaker discrimination capability \cite{pan2018two}. To address this problem, we propose to restitute the task-specific discriminative features from the IN removed information, by disentangling it into identity-relevant features and identity-irrelevant features with a dual causality loss constraint (see Figure \ref{fig:flowchart}(b)). We elaborate on the designed SNR module hereafter. 

For an SNR module, we denote the input (which is a feature map) by $F \in \mathbb{R}^{h\times w \times c}$ and the output by $\widetilde{F}^{+} \in \mathbb{R}^{h\times w \times c}$, where $h,w,c$ denote the height, width, and number of channels, respectively.




\noindent\textbf{Style Normalization Phase.} 
In SNR, we first try to reduce the domain discrepancy on the input features by performing Instance Normalization \cite{ulyanov2016instance,dumoulin2016learned,ulyanov2017improved,huang2017arbitrary} as
\begin{equation}
    \begin{aligned}
        \widetilde{F} = {\rm {IN}}(F) = \gamma  (\frac{F-\mu(F)}{\sigma(F)}) + \beta,
    \end{aligned}
\end{equation}
where $\mu(\cdot)$ and $\sigma(\cdot)$ denote the mean and standard deviation computed across spatial dimensions independently for each channel and each \emph{sample/instance}, $\gamma$, $\beta$ $\in \mathbb{R}^c$ are parameters learned from data. IN could filter out some instance-specific style information from the content. With IN taking place in the feature space, Huang \etal \cite{huang2017arbitrary} have argued and experimentally shown that IN has more profound impacts than a simple contrast normalization and it performs a form of \emph{style normalization} by normalizing feature statistics. 

\noindent\textbf{Style Restitution Phase.} IN reduces style discrepancy and boosts the generalization capability. However, with the mathematical operations being deterministic and task-irrelevant, it inevitably discards some discriminative (task-relevant) information for ReID. We propose to restitute the identity-relevant feature to the network by distilling it from the residual feature $R$. $R$ is defined as
\begin{equation}
       R = F - \widetilde{F},
    \label{eq:Residual}
\end{equation}
which denotes the difference between the original input feature $F$ and the style normalized feature $\widetilde{F}$.

Given $R$, we further disentangle it into two parts: identity-relevant feature $R^+ \in \mathbb{R}^{h\times w\times c}$ and identity-irrelevant feature $R^- \in \mathbb{R}^{h\times w\times c}$ through masking $R$ by a learned channel attention vector $\textbf{$\mathbf{a}$}=[a_1, a_2, \cdots, a_c] \in \mathbb{R}^c$:
\begin{equation}
    \begin{aligned}
        R^+(:,:,k) = & a_k R(:,:,k), \\ R^-(:,:,k) = & (1 - a_k)  R(:,:,k),
    \end{aligned}
    \label{eq:seperation}
\end{equation}
where $R(:,:,k) \in \mathbb{R}^{h\times w}$ denotes the $k^{th}$ channel of feature map $R$, $k=1,2,\cdots,c$. We expect the channel attention vector $\textbf{$\mathbf{a}$}$ to enable the adaptive distillation of the identity-relevant features for restitution, and derive it by SE-like \cite{hu2018squeeze} channel attention as
\begin{equation}
    \begin{aligned}
        \textbf{$\mathbf{a}$} = g(R) = \sigma({\rm W_2}\delta({\rm W_1} pool(R))),
    \end{aligned}
    \label{eq:se}
\end{equation}
which consists of a global average pooling layer followed by two FC layers that are parameterized by ${{\rm W_2}} \in \mathbb{R}^{(c/r) \times c}$ and ${{\rm W_1}} \in \mathbb{R}^{c \times  (c/r)}$ which are followed by ReLU activation function $\delta(\cdot)$ and sigmoid activation function $\sigma(\cdot)$, respectively. To reduce the number of parameters, a dimension reduction ratio $r$ is used and is set to 16.

By adding the distilled identity-relevant feature $R^+$ to the style normalized feature $\widetilde{F}$, we obtain the output feature $\widetilde{F}^+$ of the SNR module as
\begin{equation}
       \widetilde{F}^+ = \widetilde{F} + R^+.
    \label{eq:addrelevant}
\end{equation}

\noindent\textbf{Dual Causality Loss Constraint.}
In order to facilitate the disentanglement of  identity-relevant feature and identity-irrelevant feature, we design a dual causality loss constraint by comparing the discrimination capability of features \emph{before} and \emph{after} the restitution. As illustrated in Figure \ref{fig:flowchart}(c), the main idea is that: after restituting the identity-relevant feature $R^+$ to the normalized feature $\widetilde{F}$, the feature becomes more discriminative; On the other hand, after restituting the identity-irrelevant feature $R^-$ to the normalized feature $\widetilde{F}$, the feature should become less discriminative. We achieve this by defining a dual causality loss $\mathcal{L}_{SNR}$ which consists of \emph{clarification loss} $\mathcal{L}_{SNR}^+$ and \emph{destruction loss} $\mathcal{L}_{SNR}^-$, \ieno, $\mathcal{L}_{SNR} = \mathcal{L}_{SNR}^+ + \mathcal{L}_{SNR}^-$.

Within a mini-batch, we sample three images, \ieno, an anchor sample $a$, a positive sample $p$ that has the same identity as the anchor sample, and a negative sample $n$ that has a different identity from the anchor sample. For simplicity, we differentiate the three samples by subscript. For example, the style normalized feature of sample $a$ is denoted by $\widetilde{F}_a$. 

Intuitively, adding the identity-relevant feature $R^+$ to the normalized feature $\widetilde{F}$, which we refer to as \emph{enhanced feature} $\widetilde{F}^+ = \widetilde{F} + R^+$, results in better discrimination capability --- the sample features with same identities are closer and those with different identities are farther apart. We calculate the distances between samples on a spatially average pooled feature to avoid the distraction caused by spatial misalignment among samples (\egno, due to different poses/viewpoints).
We denote the spatially average pooled feature of $\widetilde{F}$ and $\widetilde{F}^+$ as $\widetilde{\mathbf{f}} = pool(\widetilde{F})$, $\widetilde{\mathbf{f}}^+ = pool(\widetilde{F}^+)$, respectively. The \emph{clarification loss} is thus defined as
\begin{equation}
    \begin{aligned}
        \mathcal{L}_{SNR}^+ & = Softplus( d(\widetilde{\mathbf{f}}_a^+,\widetilde{\mathbf{f}}_p^+) - d(\widetilde{\mathbf{f}}_a, \widetilde{\mathbf{f}}_p)) \\
        & + Softplus( d(\widetilde{\mathbf{f}}_a,\widetilde{\mathbf{f}}_n) - d(\widetilde{\mathbf{f}}_a^+, \widetilde{\mathbf{f}}_n^+)),
    \end{aligned}
\end{equation}
where $d(\mathbf{x},\mathbf{y})$ denotes the  distance between $\mathbf{x}$ and $\mathbf{y}$ which is defined as $d(\mathbf{x},\mathbf{y}) = 0.5 - \mathbf{x}^{\rm{T}} \mathbf{y}/ (2\lVert\mathbf{x}\rVert\lVert\mathbf{y}\lVert)$. $Softplus(\cdot) = ln(1+exp(\cdot))$ is a monotonically increasing function that aims to reduce the optimization difficulty by avoiding negative loss values.   


On the other hand, we expect that the adding of the identity-irrelevant feature $R^-$ to the normalized feature $\widetilde{F}$, which we refer to as \emph{contaminated feature} $\widetilde{F}^- = \widetilde{F} + R^-$, could decrease the discrimination capability. In comparison with the normalized feature $\widetilde{F}$ before the compensation, we expect that adding $R^-$ would push the sample features with same identities farther apart and pull those with different identities closer. We denote the spatially average pooled feature of $\widetilde{F}^-$ as $\widetilde{\mathbf{f}}^- = pool(\widetilde{F}^-)$. The \emph{destruction loss} is:
\begin{equation}
    \begin{aligned}
        \mathcal{L}_{SNR}^- & = Softplus( d(\widetilde{\mathbf{f}}_a,\widetilde{\mathbf{f}}_p) - d(\widetilde{\mathbf{f}}_a^-, \widetilde{\mathbf{f}}_p^-)) \\
        & + Softplus( d(\widetilde{\mathbf{f}}_a^-,\widetilde{\mathbf{f}}_n^-) - d(\widetilde{\mathbf{f}}_a, \widetilde{\mathbf{f}}_n)).
    \end{aligned}
\end{equation}

\subsection{Joint Training}

We use the commonly used ResNet-50 as a base ReID network and insert the proposed SNR module after each convolution block (in total four convolution blocks/stages)(see Figure \ref{fig:flowchart}(a)). We train the entire network in an end-to-end manner. The overall loss is 
\begin{equation}
    \begin{aligned}
        \mathcal{L} = \mathcal{L}_{ReID} +  &\sum_{b=1}^4 \lambda_b \mathcal{L}_{SNR}^b,
        \label{eq:total-loss}
    \end{aligned}
\end{equation}
where $\mathcal{L}_{SNR}^b$ denotes the dual causality loss for the $b^{th}$ SNR module. $\mathcal{L}_{ReID}$ denotes the widely-used ReID Loss (classification loss \cite{sun2018beyond,fu2019horizontal}, and triplet loss with batch hard mining \cite{hermans2017defense}) on the ReID feature vectors. $\lambda_b$ is a weight which controls the relative importance of the regularization at stage $b$. In considering that the features of stage 3 and 4 are more relevant to the task (high-level semantics), we experimentally set $\lambda_{3}$, $\lambda_{4}$ to 0.5, and $\lambda_1$,$\lambda_2$ to 0.1.

\section{Experiments}
In this section, we first describe the datasets and evaluation metrics in Subsection \ref{subsec:dataset}. Then, for generalizable ReID, we validate the effectiveness of SNR in Subsection \ref{subsec:ablation} and study its design choices in Subsection \ref{subsec:design}. We conduct visualization analysis in Subsection \ref{subsec:visualization}. Subsection \ref{subsec:SOTA} shows the comparisons of our schemes with the state-of-the-art approaches for both generalizable person ReID and unsupervised domain adapation ReID, respectively. In Subsection \ref{subsec:extension}, we further validate the effectiveness of applying the SNR modules to another backbone network and to cross modality (Infrared-RGB) person ReID. 

We use ResNet-50 \cite{he2016deep,almazan2018re,zhang2019DSA,luo2019bag} as our base network for both baselines and our schemes. We build a strong baseline \emph{Baseline} with some commonly used tricks integrated. 




\subsection{Datasets and Evaluation Metrics}
\label{subsec:dataset}

To evaluate the generalization ability of our approach and to be consistent with what were done in prior works for performance comparisons, we 
conduct extensive 
experiments on commonly used public ReID datasets, including 
Market1501 \cite{zheng2015scalable}, DukeMTMC-reID \cite{zheng2017unlabeled}, CUHK03 \cite{li2014deepreid}, the large-scale MSMT17 \cite{wei2018person}, and four small-scale ReID datasets of PRID \cite{hirzer2011person}, GRID \cite{loy2010time}, VIPeR \cite{gray2008viewpoint}, and i-LIDS \cite{wei2009associating}. We denote Market1501 by M, DukeMTMC-reID by Duke or D, and CUHK03 by C for simplicity. 


We follow common practices and use the cumulative matching characteristics (CMC) at Rank-1, and mean average precision (mAP) to evaluate the performance.




\begin{table*}[htbp]
  \centering
  \scriptsize
  \vspace{-6mm}
  \caption{Performance (\%) comparisons of our scheme and others to demonstrate the effectiveness of our SNR module for generalizable person ReID. The rows denote source dataset(s) for training and the columns correspond to different target datasets for testing. We mask the results of supervised ReID by gray where the testing domain has been seen in training. Due to space limitation, we only show a portion of the results here and more comparisons can be found in \textbf{Supplementary}.}
  \vspace{-5mm}
    \begin{tabular}{c|l|cccccccccccc}
    \multicolumn{1}{r}{} & \multicolumn{1}{r}{} &       &       &       &       &       &       &       &       &       &       &       &  \\
    \midrule
    \multirow{2}[2]{*}{Source} & \multicolumn{1}{c|}{\multirow{2}[2]{*}{Method}} & \multicolumn{2}{c}{Target: Market1501} & \multicolumn{2}{c}{Target: Duke} & \multicolumn{2}{c}{Target: PRID} & \multicolumn{2}{c}{Target: GRID} & \multicolumn{2}{c}{Target: VIPeR} & \multicolumn{2}{c}{Target: iLIDs} \\
          &       & mAP   & Rank-1 & mAP   & Rank-1 & mAP   & Rank-1 & mAP   & Rank-1 & mAP   & Rank-1 & mAP   & Rank-1 \\
    \midrule
    \multirow{6}[2]{*}{Market1501 (M)} & Baseline & \cellcolor[rgb]{ .906,  .902,  .902}82.8 & \cellcolor[rgb]{ .906,  .902,  .902}93.2 & 19.8  & 35.3  & 13.7  & 6.0     & 25.8  & 16.0    & 37.6  & 28.5  & 61.5  & 53.3 \\
          & Baseline-A-IN & \cellcolor[rgb]{ .906,  .902,  .902}75.3 & \cellcolor[rgb]{ .906,  .902,  .902}89.8 & 24.1  & 42.7  & 33.9  & 21.0    & 35.6  & 27.2  & \textcolor[rgb]{ .357,  .608,  .835}{38.1} & \textcolor[rgb]{ .357,  .608,  .835}{29.1} & \textcolor[rgb]{ .357,  .608,  .835}{64.2} & \textcolor[rgb]{ .357,  .608,  .835}{55.0} \\
          & Baseline-IBN & \cellcolor[rgb]{ .906,  .902,  .902}81.1 & \cellcolor[rgb]{ .906,  .902,  .902}92.2 & 21.5  & 39.2  & 19.1  & 12.0    & 27.5  & 19.2  & 32.1  & 23.4  & 58.3  & 48.3 \\
          & Baseline-A-SN & \cellcolor[rgb]{ .906,  .902,  .902}\textcolor[rgb]{ .357,  .608,  .835}{83.2} & \cellcolor[rgb]{ .906,  .902,  .902}\textcolor[rgb]{ .357,  .608,  .835}{93.9} & 20.1  & 38.0    & \textcolor[rgb]{ .357,  .608,  .835}{35.4} & \textcolor[rgb]{ .357,  .608,  .835}{25.0} & 29.0    & 22.0    & 32.2  & 23.4  & 53.4  & 43.3 \\
          & Baseline-IN & \cellcolor[rgb]{ .906,  .902,  .902}79.5 & \cellcolor[rgb]{ .906,  .902,  .902}90.9 & \textcolor[rgb]{ .357,  .608,  .835}{25.1} & \textcolor[rgb]{ .357,  .608,  .835}{44.9} & 35.0    & \textcolor[rgb]{ .357,  .608,  .835}{25.0} & \textcolor[rgb]{ .357,  .608,  .835}{35.7} & \textcolor[rgb]{ .357,  .608,  .835}{27.8} & 35.1  & 27.5  & 64.0    & 54.2 \\
          & \textbf{Baseline-SNR (Ours)}  & \cellcolor[rgb]{ .906,  .902,  .902}\textcolor[rgb]{ 1,  0,  0}{\textbf{84.7}} & \cellcolor[rgb]{ .906,  .902,  .902}\textcolor[rgb]{ 1,  0,  0}{\textbf{94.4}} & \textcolor[rgb]{ 1,  0,  0}{\textbf{33.6}} & \textcolor[rgb]{ 1,  0,  0}{\textbf{55.1}} & \textcolor[rgb]{ 1,  0,  0}{\textbf{42.2}} & \textcolor[rgb]{ 1,  0,  0}{\textbf{30.0}} & \textcolor[rgb]{ 1,  0,  0}{\textbf{36.7}} & \textcolor[rgb]{ 1,  0,  0}{\textbf{29.0}} & \textcolor[rgb]{ 1,  0,  0}{\textbf{42.3}} & \textcolor[rgb]{ 1,  0,  0}{\textbf{32.3}} & \textcolor[rgb]{ 1,  0,  0}{\textbf{65.6}} & \textcolor[rgb]{ 1,  0,  0}{\textbf{56.7}} \\
    \midrule
    \multirow{6}[2]{*}{Duke (D)} & Baseline & 21.8  & 48.3  & \cellcolor[rgb]{ .906,  .902,  .902}71.2 & \cellcolor[rgb]{ .906,  .902,  .902}83.4 & 15.7  & 11.0    & 14.5  & 8.8   & \textcolor[rgb]{ .357,  .608,  .835}{37.0} & 26.9  & 68.3  & 58.3 \\
          & Baseline-A-IN & 26.5  & 56.0    & \cellcolor[rgb]{ .906,  .902,  .902}64.5 & \cellcolor[rgb]{ .906,  .902,  .902}78.9 & 38.6  & 29.0    & 19.6  & \textcolor[rgb]{ .357,  .608,  .835}{13.6} & 35.1  & \textcolor[rgb]{ .357,  .608,  .835}{27.2} & 67.4  & 56.7 \\
          & Baseline-IBN & 24.6  & 52.5  & \cellcolor[rgb]{ .906,  .902,  .902}69.5 & \cellcolor[rgb]{ .906,  .902,  .902}81.4 & 27.4  & 19.0    & 19.9  & 12.0    & 32.8  & 23.4  & 63.5  & 61.7 \\
          & Baseline-A-SN & 25.3  & 55.0    & \cellcolor[rgb]{ .906,  .902,  .902}\textcolor[rgb]{ 1,  0,  0}{\textbf{73.0}} & \cellcolor[rgb]{ .906,  .902,  .902}\textcolor[rgb]{1,  0,  0}{\textbf{85.9}} & \textcolor[rgb]{ .357,  .608,  .835}{41.4} & \textcolor[rgb]{ .357,  .608,  .835}{32.0} & 18.8  & 12.8  & 31.3  & 24.1  & 64.8  & 63.3 \\
          & Baseline-IN & \textcolor[rgb]{ .357,  .608,  .835}{27.2} & \textcolor[rgb]{ .357,  .608,  .835}{58.5} & \cellcolor[rgb]{ .906,  .902,  .902}68.9 & \cellcolor[rgb]{ .906,  .902,  .902}80.4 & 40.5  & 27.0    & \textcolor[rgb]{ .357,  .608,  .835}{20.3} & 13.2  & 34.6  & 26.3  & \textcolor[rgb]{ .357,  .608,  .835}{70.6} & \textcolor[rgb]{ .357,  .608,  .835}{65.0} \\
          & \textbf{Baseline-SNR (Ours)}  & \textcolor[rgb]{ 1,  0,  0}{\textbf{33.9}} & \textcolor[rgb]{ 1,  0,  0}{\textbf{66.7}} & \cellcolor[rgb]{ .906,  .902,  .902}\textcolor[rgb]{ .357,  .608,  .835}{{72.9}} & \cellcolor[rgb]{ .906,  .902,  .902}\textcolor[rgb]{ .357,  .608,  .835}{{84.4}} & \textcolor[rgb]{ 1,  0,  0}{\textbf{45.4}} & \textcolor[rgb]{ 1,  0,  0}{\textbf{35.0}} & \textcolor[rgb]{ 1,  0,  0}{\textbf{35.3}} & \textcolor[rgb]{ 1,  0,  0}{\textbf{26.0}} & \textcolor[rgb]{ 1,  0,  0}{\textbf{41.2}} & \textcolor[rgb]{ 1,  0,  0}{\textbf{32.6}} & \textcolor[rgb]{ 1,  0,  0}{\textbf{79.3}} & \textcolor[rgb]{ 1,  0,  0}{\textbf{68.7}} \\
    \midrule

    \multirow{2}[0]{*}{\begin{tabular}[c]{@{}c@{}}M + D + CUHK03\\ + MSMT17\end{tabular}} & Baseline & \cellcolor[rgb]{ .906,  .902,  .902}\textcolor[rgb]{ .357,  .608,  .835}{72.4} & \cellcolor[rgb]{ .906,  .902,  .902}\textcolor[rgb]{ .357,  .608,  .835}{88.7} & \cellcolor[rgb]{ .906,  .902,  .902}\textcolor[rgb]{ .357,  .608,  .835}{70.1}  & \cellcolor[rgb]{ .906,  .902,  .902}\textcolor[rgb]{ .357,  .608,  .835}{83.8} & \textcolor[rgb]{ .357,  .608,  .835}{39.0} & \textcolor[rgb]{ .357,  .608,  .835}{28.0}    & \textcolor[rgb]{ .357,  .608,  .835}{29.6}    & \textcolor[rgb]{ .357,  .608,  .835}{20.8}  & \textcolor[rgb]{ .357,  .608,  .835}{52.1}  & \textcolor[rgb]{ .357,  .608,  .835}{41.5}  & \textcolor[rgb]{ .357,  .608,  .835}{89.0}  & \textcolor[rgb]{ .357,  .608,  .835}{85.0} \\
	
	& \textbf{Baseline-SNR (Ours)} &\cellcolor[rgb]{ .906,  .902,  .902} \textcolor[rgb]{ 1,  0,  0}{\textbf{82.3}} & \cellcolor[rgb]{ .906,  .902,  .902} \textcolor[rgb]{ 1,  0,  0}{\textbf{93.4}} & \cellcolor[rgb]{ .906,  .902,  .902} \textcolor[rgb]{ 1,  0,  0}{\textbf{73.2}} & \cellcolor[rgb]{ .906,  .902,  .902} \textcolor[rgb]{ 1,  0,  0}{\textbf{85.5}} & \cellcolor[rgb]{ 1,  1,  1}\textcolor[rgb]{ 1,  0,  0}{\textbf{60.0}} & \cellcolor[rgb]{ 1,  1,  1}\textcolor[rgb]{ 1,  0,  0}{\textbf{49.0}} & \cellcolor[rgb]{ 1,  1,  1}\textcolor[rgb]{ 1,  0,  0}{\textbf{41.3}} & \cellcolor[rgb]{ 1,  1,  1}\textcolor[rgb]{ 1,  0,  0}{\textbf{30.4}} & \cellcolor[rgb]{ 1,  1,  1}\textcolor[rgb]{ 1,  0,  0}{\textbf{65.0}} & \cellcolor[rgb]{ 1,  1,  1}\textcolor[rgb]{ 1,  0,  0}{\textbf{55.1}} & \cellcolor[rgb]{ 1,  1,  1}\textcolor[rgb]{ 1,  0,  0}{\textbf{91.9}} & \cellcolor[rgb]{ 1,  1,  1}\textcolor[rgb]{ 1,  0,  0}{\textbf{87.0}} \\
         
    \bottomrule
    \end{tabular}%
    \vspace{-3mm}
  \label{tab:ablationstudy}%
\end{table*}%

\subsection{Ablation Study}
\label{subsec:ablation}

We perform comprehensive ablation studies to demonstrate the effectiveness of the SNR module and its dual causality loss constraint. We mimic the real-world scenario for generalizable person ReID, where a model is trained on some source dataset(s) A while tested on previously unseen dataset B. We denote this as A$\rightarrow $B. We have several experimental settings to evaluate the generalization capability, \egno, Market1501$\rightarrow$Duke and others, Duke$\rightarrow$Market1501 and others, M+D+C+MSMT17$\rightarrow$others. Our settings cover both single source dataset for training and multiple source datasets for training. 




\noindent\textbf{Effectiveness of Our SNR.}
Here we compare several schemes. \textbf{\emph{Baseline}}: a strong baseline based on ResNet-50. \textbf{\emph{Baseline-A-IN}}: a naive model where we replace all the Batch Normalization(BN) \cite{ioffe2015batch} layers in \emph{Baseline} by Instance Normalization(IN). \textbf{\emph{Baseline-IBN}}: Similar to IBN-Net (IBN-b)~\cite{pan2018two} and OSNet \cite{zhou2019omni}, we add IN only to the last layers of Conv1 and Conv2 blocks of \emph{Baseline} respectively. \textbf{\emph{Baseline-A-SN}}: a model where we replace all the BN layers in \emph{Baseline} by Switchable Normalization (SN). SN \cite{luo2018differentiable} can be regarded as an adaptive ensemble version of normalization techniques of IN, BN, and LN (Layer Normalization) \cite{ba2016layer}. \textbf{\emph{Baseline-IN}}: four IN layers are added after the first four convolutional blocks/stages of \emph{Baseline} respectively. \textbf{\emph{Baseline-SNR}}: our final scheme where four SNR modules are added after the first four convolutional blocks/stages of \emph{Baseline} respectively (see Figure~\ref{fig:flowchart}(a)). We also refer to it as \textbf{\emph{SNR}} for simplicity. Table \ref{tab:ablationstudy} shows the results. We have the following observations/conclusions:


\noindent\textbf{1)} {\emph{Baseline-A-IN}} improves \emph{Baseline} by \textbf{4.3\%} in mAP for Market1501$\rightarrow$Duke, and \textbf{4.7\%} in mAP for Duke$\rightarrow$Market1501. Other IN-related baselines also bring gains, which demonstrates the effectiveness of IN for improving the generalization capability for ReID. But, IN also inevitably discards some discriminative (identity-relevant) information and we can see it clearly decreases the performance of \emph{Baseline-A-IN}, \emph{Baseline-IBN} and \emph{Baseline-IN} for the same-domain ReID (\egno, Market1501$\rightarrow$Market1501). \emph{Baseline-A-SN} learns the combination weights of IN, BN, and LN in the training dataset and thus has superior performance in the same domain, but it does not have dedicated design for boosting the generalization capability. 

\noindent\textbf{2)} Thanks to the compensation of the identity-relevant information through the proposed \emph{restitution step}, our final scheme {\emph{Baseline-SNR}} achieves superior generalization capability, which significantly outperforms all the baseline schemes. In particular, \emph{Baseline-SNR} outperforms \emph{Baseline-IN} by \textbf{8.5\%}, \textbf{6.7\%}, \textbf{15.0\%} in mAP for M$\rightarrow$D, D$\rightarrow$M, and D$\rightarrow$GRID, respectively.



\noindent\textbf{3)} The generalization performance on previously unseen target domain increases consistently as the number of source datasets increases. When all the four source datasets are used (the large-scale MSMT17 \cite{wei2018person} also included), we have a very strong baseline (\ieno, 52.1\% in mAP on VIPeR dataset vs. 37.6\% when Market1501 alone is used as source). Interestingly, our method still significantly outperforms the strong baseline \emph{Baseline}, even by \textbf{21.0\%} in mAP on PRID dataset, demonstrating SNR's effectiveness.


\begin{table*}[t]\centering\vspace{-6mm}
    \caption{Effectiveness of dual causality loss constraint (a), and study on design choices of SNR (b) and (c).}
    \vspace{-2mm}
	\captionsetup[subfloat]{captionskip=2pt}
	\captionsetup[subffloat]{justification=centering}
	\subfloat[Study on the dual causality loss constraint.\label{tab:Loss}]{
		\tablestyle{3.33pt}{1.05}
		\begin{tabular}{lcccc}
            \toprule
            \multicolumn{1}{c}{\multirow{2}[1]{*}{Method}} & \multicolumn{2}{c}{M$\longrightarrow$D} & \multicolumn{2}{c}{D$\longrightarrow$M} \\
                  & mAP   & Rank-1 & mAP   & Rank-1 \\
            \midrule
            Baseline & 19.8  & 35.3  & 21.8  & 48.3 \\
            SNR w/o $\mathcal{L}_{SNR}$
            & 26.1  & 45.0  & 29.2  & 57.4 \\
            SNR w/o $\mathcal{L}_{SNR}^+$ & {28.8} & {48.9} & 30.2  & {59.8} \\
            SNR w/o $\mathcal{L}_{SNR}^-$ & 28.0  & 48.1  & {30.3} & 59.1 \\
            \textbf{SNR}  & \textcolor[rgb]{ 1,  0,  0}{\textbf{33.6}} & \textcolor[rgb]{ 1,  0,  0}{\textbf{55.1}} & \textcolor[rgb]{ 1,  0,  0}{\textbf{33.9}} & \textcolor[rgb]{ 1,  0,  0}{\textbf{66.7}} \\
            \bottomrule
            \end{tabular}}\hspace{3mm}
	\subfloat[Study on which stage to add SNR. \label{tab:stage}]{
		\tablestyle{3.33pt}{0.92}
		\begin{tabular}{lcccc}
                \toprule
                \multicolumn{1}{c}{\multirow{2}[2]{*}{Method}} & \multicolumn{2}{c}{M$\longrightarrow$D} & \multicolumn{2}{c}{D$\longrightarrow$M} \\
                      & mAP   & Rank-1 & mAP   & Rank-1 \\
                \midrule
                Baseline & 19.8  & 35.3  & 21.8  & 48.3 \\
                stage-1 & 23.7  & 42.8  & 27.6  & 57.7 \\
                stage-2 & 24.0    & 44.4  & 28.6  & 58.8 \\
                stage-3 & {26.4} & {46.3} & {29.5} & {60.7} \\
                stage-4 & 26.2  & 45.8  & 29.4  & 59.7 \\
                \textbf{stages-all}  & \textcolor[rgb]{ 1,  0,  0}{\textbf{33.6}} & \textcolor[rgb]{ 1,  0,  0}{\textbf{55.1}} & \textcolor[rgb]{ 1,  0,  0}{\textbf{33.9}} & \textcolor[rgb]{ 1,  0,  0}{\textbf{66.7}} \\
                \bottomrule
                \end{tabular}}\hspace{3mm}
	\subfloat[Disentanglement designs in SNR.\label{tab:so}]{
		\tablestyle{3.33pt}{1.22}
		\begin{tabular}{lcccc}
                \toprule
                \multicolumn{1}{c}{\multirow{2}[1]{*}{Method}} & \multicolumn{2}{c}{M$\longrightarrow$D} & \multicolumn{2}{c}{D$\longrightarrow$M} \\
                      & mAP   & Rank-1 & mAP   & Rank-1 \\
                \midrule
                Baseline & 19.8  & 35.3  & 21.8  & 48.3 \\
                SNR$_{conv}$ & 29.7  & 51.1  & 29.4  & 61.7 \\
                SNR$_{g(\cdot)^2}$ & {31.2} & {52.9} & {31.0} & {63.8} \\
                \textbf{SNR}  & \textcolor[rgb]{ 1,  0,  0}{\textbf{33.6}} & \textcolor[rgb]{ 1,  0,  0}{\textbf{55.1}} & \textcolor[rgb]{ 1,  0,  0}{\textbf{33.9}} & \textcolor[rgb]{ 1,  0,  0}{\textbf{66.7}} \\
                \bottomrule
            \end{tabular}
	}
	\vspace{-8mm}
	\label{tab:ablations}
\end{table*}

\noindent\textbf{4)} The performance of different schemes with respects to PRID/GRID varies greatly and the mAPs are all relatively low, which is caused by the large style discrepancy between PRID/GRID and other datasets. For such challenging cases, our scheme still outperforms \emph{Baseline-IN} significantly by \textbf{7.2\%} and \textbf{4.9\%} in mAP for M$\rightarrow$PRID and D$\rightarrow$PRID, respectively.




\noindent\textbf{5)} For supervised ReID (masked by gray), our scheme also clearly outperforms \emph{Baseline} by \textbf{1.9\%} and \textbf{1.7\%} in mAP for M$\rightarrow$M and D$\rightarrow$D, respectively. That is because there is also style discrepancy within the source domain.

\noindent\textbf{Influence of Dual Causality Loss Constraint.}
We study the effectiveness of the proposed dual causality loss $\mathcal{L}_{SNR}$ which consists of \emph{clarification loss} $\mathcal{L}_{SNR}^+$ and \emph{destruction loss} $\mathcal{L}_{SNR}^-$. Table \ref{tab:Loss} shows the results. Our final scheme \emph{SNR} with the dual causality loss $\mathcal{L}_{SNR}$ outperforms that without such constraints (\ieno, scheme \emph{SNR w/o $\mathcal{L}_{SNR}$}) by \textbf{7.5\%} and \textbf{4.7\%} in mAP for M$\rightarrow$D and D$\rightarrow$M, respectively. Such constraints facilitate the disentanglement of identity-relevant/identity-irrelevant features. In addition, both the clarification loss $\mathcal{L}_{SNR}^+$ and the destruction loss $\mathcal{L}_{SNR}^-$, are vital to SNR and they are complementary and jointly contribute to a superior performance. 



\noindent\textbf{Complexity.} The model size of our final scheme \emph{SNR} is very similar to that of \emph{Baseline} (24.74 M vs. 24.56 M).


\subsection{Design Choices of SNR}
\label{subsec:design}

\noindent\textbf{Which Stage to Add SNR?} We compare the cases of adding a single SNR module to a different convolutional block/stage, and to all the four stages (\ie, stage-1 $\sim$ 4) of the ResNet-50 (see Figure \ref{fig:flowchart}(a)). The module is added after the last layer of a convolutional block/stage. As Table \ref{tab:stage} shows, in comparison with \emph{Baseline}, the improvement from adding SNR is significant on stage-3 and stage-4 and is a little smaller on stage-1 and stage-2. When SNR is added to all the four stages, we achieve the best performance.




\noindent\textbf{Influence of Disentanglement Design.}
In our SNR module, as described in (\ref{eq:seperation})(\ref{eq:se}) of Subsection \ref{subsec:SNR}, we use $g(\cdot)$, and its complementary one $1-g(\cdot)$ as masks to extract identity-relevant feature $R^+$ and identity-irrelevant feature $R^-$ from the residual feature $R$. Here, we study the influence of different disentanglement designs within SNR. \textbf{\emph{SNR${_{conv}}$}}: we disentangle the residual feature $R$ through 1$\times$1 convolutional layer followed by non-liner ReLU activation, \ieno, $R^+ = ReLU(W^+ R)$, $R^- = ReLU(W^- R)$. \textbf{\emph{SNR${_{g(\cdot)^2}}$}}: we use two unshared gates $g(\cdot)^+$, $g(\cdot)^-$ to obtain $R^+$ and $R^-$ respectively. Table \ref{tab:so} shows the results. We observe that (1) ours outperforms \textbf{\emph{SNR${_{conv}}$}} by \textbf{3.9\%} and \textbf{4.5\%} in mAP for M$\rightarrow$D and D$\rightarrow$M, respectively, demonstrating the benefit of content-adaptive design; (2) ours outperforms \textbf{\emph{SNR$_{g(\cdot)^2}$}} by \textbf{2.4\%/2.9\%} in mAP on the unseen target Duke/Market1501, demonstrating the benefit of the design which encourages interaction between $R^+$ and $R^-$.

\begin{figure}
  \centerline{\includegraphics[width=1.0\linewidth]{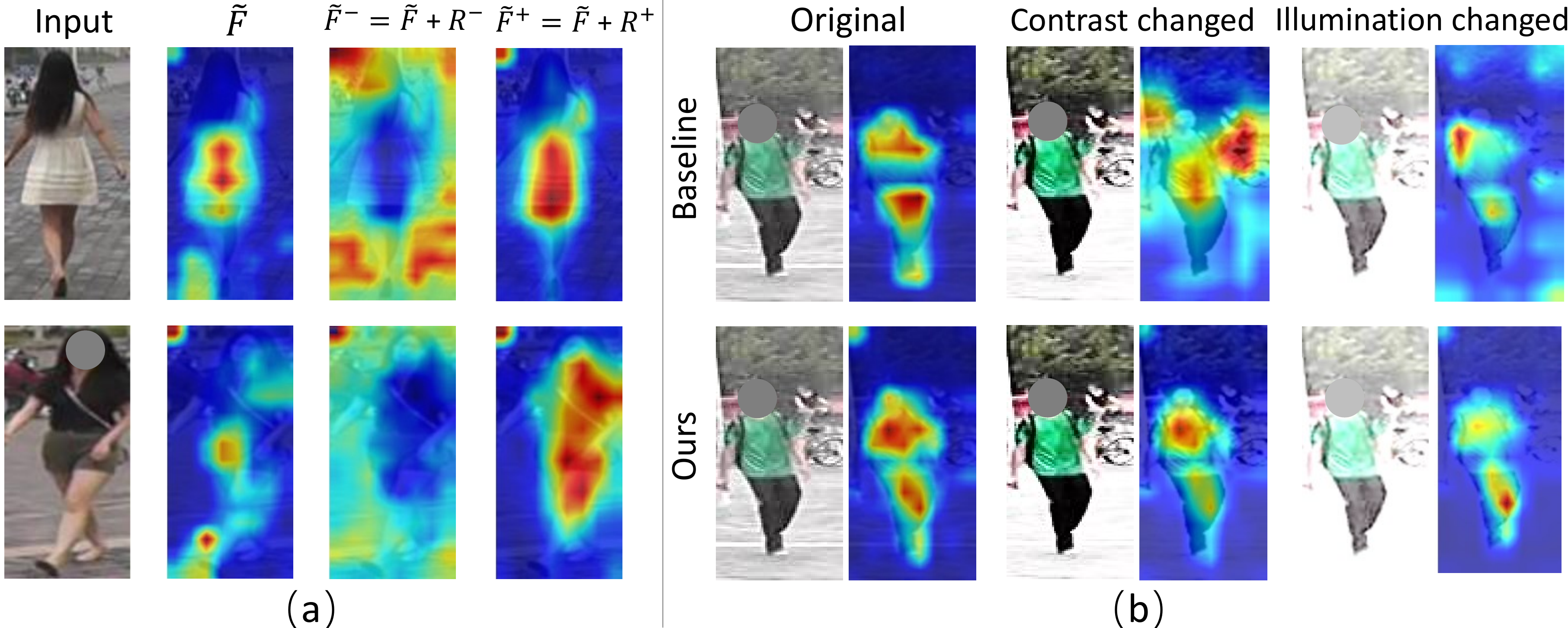}}
  \vspace{-3mm}
  \caption{(a) Activation maps of different features within an SNR module (SNR 3). They show SNR can disentangle the identity-relevant/irrelevant features well. (b) Activation maps of our scheme (bottom) and the strong baseline  \emph{Baseline} (top) corresponding to images of varied styles. Our maps are more consistent/invariant to style variants.}
\label{fig:vis_ftp}
 \vspace{-3mm}
\end{figure}

\vspace{-1mm}
\subsection{Visualization}
\label{subsec:visualization}
\vspace{-1mm}
\noindent\textbf{Feature Map Visualization.} To better understand how an SNR module works, we visualize the intermediate feature maps of the third SNR module (SNR 3). Following \cite{zhou2019omni,zheng2011person}, we get each activation map by summarizing the feature maps along channels followed by a spatial $\ell_2$ normalization.


Figure \ref{fig:vis_ftp}(a) shows the activation maps of normalized feature $\widetilde{F}$, enhanced feature $\widetilde{F}^+=\widetilde{F}+R^+$, and contaminated feature $\widetilde{F}^-=\widetilde{F}+R^-$, respectively. We see that after adding the identity-irrelevant feature $R^{-}$, the contaminated feature $\widetilde{F}^-$ has high response mainly on background. In contrast, the enhanced feature $\widetilde{F}^+$ with the restitution of identity-relevant feature $R^{+}$ has high responses on regions of the human body, better capturing discriminative regions. 

Moreover, in Figure \ref{fig:vis_ftp}(b), we further compare the activation maps $\widetilde{F}^+$ of our scheme and those of the strong baseline scheme \emph{Baseline} by varying the styles of input images (\egno, contrast, illumination, saturation). We can see that, for the images with different styles, the activation maps of our scheme are more consistent/invariant than those of \emph{Baseline}. In contrast, the activation maps of \emph{Baseline} are more disorganized and are easily affected by style variants. These indicate our scheme is more robust to style variations.

\begin{figure}
  \centerline{\includegraphics[width=1.0\linewidth]{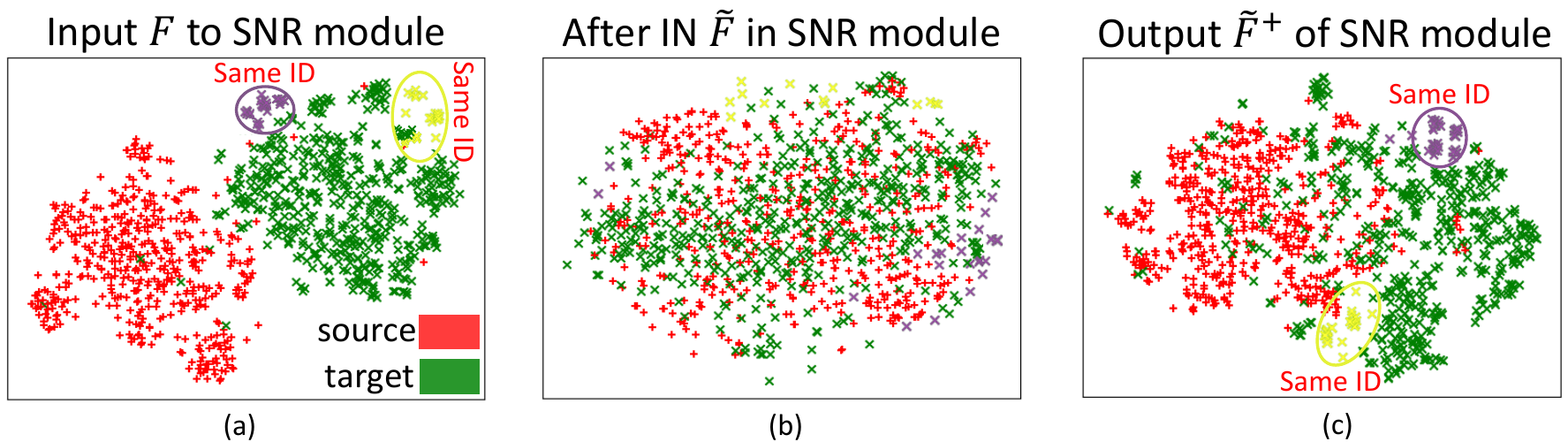}}
  \vspace{-3mm}
  \caption{Visualization of distributions of intermediate features before/within/after the SNR module using the tool of t-SNE \cite{maaten2008visualizing}. `Red'/`green' nodes: samples from source dataset Market1501/unseen target dataset Duke.}
\label{fig:vis_tSNE_main}
 \vspace{-3mm}
\end{figure}

\noindent\textbf{Visualization of Feature Distributions.} 
In Figure \ref{fig:vis_tSNE_main}, we visualize the distribution of the features from the 3$^{rd}$ SNR module of our network using t-SNE \cite{maaten2008visualizing}. They denote the distributions of features for (a) input $F$, (b) style normalized feature $\widetilde{F}$, and (c) output $\widetilde{F}^+$ of the SNR module. We observe that, (a) before SNR, the extracted features from two datasets (`red': source training dataset Market1501; `green': unseen target dataset Duke) are largely separately distributed and have an obvious \emph{domain gap}. (b) Within the SNR module, after IN, this \emph{domain gap} has been eliminated. But the samples of the same identity (`yellow' and `purple' colored nodes denote two identities respectively) become dispersive. (c) After the restitution of identity-relevant features, not only has the domain gap of feature distributions been shrunk, but also the feature distribution of samples with same identity become more compact than that in (b).

\begin{table*}[htbp]
  \centering
  \vspace{-6.5mm}
  \scriptsize
  \caption{Performance (\%) comparisons with the state-of-the-art approaches for the Domain Generalizable Person ReID (top rows) and the Unsupervised Domain Adaptation Person ReID (bottom rows), respectively. ``(U)" denotes ``unlabeled". We mask the schemes that use our \emph{Baseline} and those that use our SNR modules by gray, which provides fair comparison.}
  \vspace{-2mm}
    \begin{tabular}{c|cccccccc}
    \toprule
    \multicolumn{2}{c}{\multirow{2}[3]{*}{Method}} & \multirow{2}[3]{*}{Venue} & \multirow{2}[3]{*}{Source} & \multicolumn{2}{c}{Target: Duke} & \multirow{2}[3]{*}{Source} & \multicolumn{2}{c}{Taeget: Market1501} \\
\cmidrule{5-6}\cmidrule{8-9}    \multicolumn{2}{c}{} &       &       & mAP   & Rank-1    &       & mAP   & Rank-1 \\
    \midrule
    \multicolumn{1}{c|}{\multirow{8}[3]{*}{\begin{tabular}[c]{@{}c@{}} Domain\\Generalization \\(w/o using  \\ target data)\end{tabular}}}
          & OSNet-IBN \cite{zhou2019omni} & ICCV'19 & Market1501 & \textcolor[rgb]{ .357,  .608,  .835}{26.7}  & \textcolor[rgb]{ .357,  .608,  .835}{48.5}  & Duke  & \textcolor[rgb]{ .357,  .608,  .835}{26.1}  & \textcolor[rgb]{ .357,  .608,  .835}{57.7} \\
          & \cellcolor[rgb]{ .906,  .902,  .902}Baseline & \cellcolor[rgb]{ .906,  .902,  .902}This work & \cellcolor[rgb]{ .906,  .902,  .902}Market1501 & \cellcolor[rgb]{ .906,  .902,  .902}19.8 & \cellcolor[rgb]{ .906,  .902,  .902}35.3 & \cellcolor[rgb]{ .906,  .902,  .902}Duke & \cellcolor[rgb]{ .906,  .902,  .902}21.8 & \cellcolor[rgb]{ .906,  .902,  .902}48.3 \\
          
        & \cellcolor[rgb]{ .906,  .902,  .902}Baseline-IBN \cite{jia2019frustratingly} & \cellcolor[rgb]{ .906,  .902,  .902}BMVC'19 & \cellcolor[rgb]{ .906,  .902,  .902}Market1501 & \cellcolor[rgb]{ .906,  .902,  .902}21.5 & \cellcolor[rgb]{ .906,  .902,  .902}39.2 & \cellcolor[rgb]{ .906,  .902,  .902}Duke & \cellcolor[rgb]{ .906,  .902,  .902}24.6 & \cellcolor[rgb]{ .906,  .902,  .902}52.5 \\

          & \cellcolor[rgb]{ .906,  .902,  .902}\textbf{SNR(Ours)} & \cellcolor[rgb]{ .906,  .902,  .902}This work & \cellcolor[rgb]{ .906,  .902,  .902}Market1501 & \cellcolor[rgb]{ .906,  .902,  .902}\textcolor[rgb]{ 1,  0,  0}{\textbf{33.6}} & \cellcolor[rgb]{ .906,  .902,  .902}\textcolor[rgb]{ 1,  0,  0}{\textbf{55.1}} & \cellcolor[rgb]{ .906,  .902,  .902}Duke & \cellcolor[rgb]{ .906,  .902,  .902}\textcolor[rgb]{ 1,  0,  0}{\textbf{33.9}} & \cellcolor[rgb]{ .906,  .902,  .902}\textcolor[rgb]{ 1,  0,  0}{\textbf{66.7}} \\
\cmidrule{2-9}          & StrongBaseline \cite{kumar2019fairest} & ArXiv'19 & MSMT17 & {43.3}  & {64.5}  & MSMT17 & {36.6}  & {64.8} \\

          & OSNet-IBN \cite{zhou2019omni} & ICCV'19 & MSMT17 & \textcolor[rgb]{ .357,  .608,  .835}{45.6}  & \textcolor[rgb]{ .357,  .608,  .835}{67.4}  & MSMT17  & \textcolor[rgb]{ .357,  .608,  .835}{37.2}  & \textcolor[rgb]{ .357,  .608,  .835}{66.5} \\

          & \cellcolor[rgb]{ .906,  .902,  .902}Baseline & \cellcolor[rgb]{ .906,  .902,  .902}This work & \cellcolor[rgb]{ .906,  .902,  .902}MSMT17 & \cellcolor[rgb]{ .906,  .902,  .902}{39.1} & \cellcolor[rgb]{ .906,  .902,  .902}{60.4} & \cellcolor[rgb]{ .906,  .902,  .902}MSMT17 & \cellcolor[rgb]{ .906,  .902,  .902}{33.8} & \cellcolor[rgb]{ .906,  .902,  .902}{59.9} \\
          & \cellcolor[rgb]{ .906,  .902,  .902}\textbf{SNR(Ours)} & \cellcolor[rgb]{ .906,  .902,  .902}This work & \cellcolor[rgb]{ .906,  .902,  .902}MSMT17 & \cellcolor[rgb]{ .906,  .902,  .902}\textcolor[rgb]{ 1,  0,  0}{\textbf{50.0}} & \cellcolor[rgb]{ .906,  .902,  .902}\textcolor[rgb]{ 1,  0,  0}{\textbf{69.2}} & \cellcolor[rgb]{ .906,  .902,  .902}MSMT17 & \cellcolor[rgb]{ .906,  .902,  .902}\textcolor[rgb]{ 1,  0,  0}{\textbf{41.4}} & \cellcolor[rgb]{ .906,  .902,  .902}\textcolor[rgb]{ 1,  0,  0}{\textbf{70.1}} \\
    \midrule
    \midrule
    \multicolumn{1}{c|}{\multirow{11}[4]{*}{\begin{tabular}[c]{@{}c@{}}Unsupervised\\ Domain\\Adaptation \\(using unlabeled \\ target data)\end{tabular}}} 
          & ATNet \cite{liu2019adaptive} & CVPR'19 & Market1501 + Duke (U) & 24.9  & 45.1  & Duke + Market1501 (U) & 25.6  & 55.7 \\
          & CamStyle \cite{zhong2018camstyle} & TIP'19 & Market1501 + Duke (U) & 25.1  & 48.4  & Duke + Market1501 (U) & 27.4  & 58.8 \\
          & ARN \cite{li2018adaptation} & CVPRW'19 & Market1501 + Duke (U) & 33.4  & 60.2  & Duke + Market1501 (U) & 39.4  & 70.3 \\
          & ECN \cite{zhong2019invariance} & CVPR'19 & Market1501 + Duke (U) & 40.4  & 63.3  & Duke + Market1501 (U) & 43.0    & 75.1 \\
          & PAST \cite{zhang2019self} & ICCV'19 & Market1501 + Duke (U) & \textcolor[rgb]{ .357,  .608,  .835}{54.3} & 72.4  & Duke + Market1501 (U) & 54.6  & 78.4 \\
          & SSG \cite{Fu2018SelfsimilarityGA} & ICCV'19 & Market1501 + Duke (U) & 53.4  & \textcolor[rgb]{ .357,  .608,  .835}{73.0} & Duke + Market1501 (U) & \textcolor[rgb]{ .357,  .608,  .835}{58.3} & \textcolor[rgb]{ .357,  .608,  .835}{80.0} \\
          & \cellcolor[rgb]{ .906,  .902,  .902}Baseline+MAR \cite{yu2019unsupervised} & \cellcolor[rgb]{ .906,  .902,  .902}This work & \cellcolor[rgb]{ .906,  .902,  .902}Market1501 + Duke (U) & \cellcolor[rgb]{ .906,  .902,  .902}{35.2} & \cellcolor[rgb]{ .906,  .902,  .902}{56.5} & \cellcolor[rgb]{ .906,  .902,  .902}Duke + Market1501 (U) & \cellcolor[rgb]{ .906,  .902,  .902}{37.2} & \cellcolor[rgb]{ .906,  .902,  .902}{62.4} \\
          & \cellcolor[rgb]{ .906,  .902,  .902}\textbf{SNR(Ours)}+MAR \cite{yu2019unsupervised} & \cellcolor[rgb]{ .906,  .902,  .902}This work & \cellcolor[rgb]{ .906,  .902,  .902}Market1501 + Duke (U) & \cellcolor[rgb]{ .906,  .902,  .902}\textcolor[rgb]{ 1,  0,  0}{\textbf{58.1}} & \cellcolor[rgb]{ .906,  .902,  .902}\textcolor[rgb]{ 1,  0,  0}{\textbf{76.3}} & \cellcolor[rgb]{ .906,  .902,  .902}Duke + Market1501 (U) & \cellcolor[rgb]{ .906,  .902,  .902}\textcolor[rgb]{ 1,  0,  0}{\textbf{61.7}} & \cellcolor[rgb]{ .906,  .902,  .902}\textcolor[rgb]{ 1,  0,  0}{\textbf{82.8}} \\
\cmidrule{2-9}          & MAR \cite{yu2019unsupervised} & CVPR'19 & MSMT17 + Duke (U) & 48.0    & 67.1  & MSMT17 + Market1501 (U) & 40.0    & 67.7 \\
          & PAUL \cite{yang2019patch} & CVPR'19 & MSMT17 + Duke (U) & \textcolor[rgb]{ .357,  .608,  .835}{53.2}  & \textcolor[rgb]{ .357,  .608,  .835}{72.0}    & MSMT17 + Market1501 (U) & \textcolor[rgb]{ .357,  .608,  .835}{40.1}  & \textcolor[rgb]{ .357,  .608,  .835}{68.5} \\
          & \cellcolor[rgb]{ .906,  .902,  .902}Baseline+MAR \cite{yu2019unsupervised} & \cellcolor[rgb]{ .906,  .902,  .902}This work & \cellcolor[rgb]{ .906,  .902,  .902}MSMT17 + Duke (U) & \cellcolor[rgb]{ .906,  .902,  .902}{46.2} & \cellcolor[rgb]{ .906,  .902,  .902}{66.3} & \cellcolor[rgb]{ .906,  .902,  .902}MSMT17 + Market1501 (U) & \cellcolor[rgb]{ .906,  .902,  .902}{39.4} & \cellcolor[rgb]{ .906,  .902,  .902}{66.9} \\
          & \cellcolor[rgb]{ .906,  .902,  .902}\textbf{SNR(Ours)} + MAR \cite{yu2019unsupervised} & \cellcolor[rgb]{ .906,  .902,  .902}This work & \cellcolor[rgb]{ .906,  .902,  .902}MSMT17 + Duke (U) & \cellcolor[rgb]{ .906,  .902,  .902}\textcolor[rgb]{ 1,  0,  0}{\textbf{61.6}} & \cellcolor[rgb]{ .906,  .902,  .902}\textcolor[rgb]{ 1,  0,  0}{\textbf{78.2}} & \cellcolor[rgb]{ .906,  .902,  .902}MSMT17 + Market1501 (U) & \cellcolor[rgb]{ .906,  .902,  .902}\textcolor[rgb]{ 1,  0,  0}{\textbf{65.9}} & \cellcolor[rgb]{ .906,  .902,  .902}\textcolor[rgb]{ 1,  0,  0}{\textbf{85.5}} \\
    \bottomrule
    \end{tabular}%
  \label{tab:STO}%
  \vspace{-3.5mm}
\end{table*}%

\vspace{-1mm}
\subsection{Comparison with State-of-the-Arts}
\label{subsec:SOTA}

Thanks to the capability of reducing style discrepancy and restitution of identity-relevant features, our proposed SNR module can enhance the generalization ability and maintain the discrimintive ability of ReID networks. It can be used for generalizable person ReID, \ieno, domain generalization (DG), and can also be used to build the backbone networks for unsupervised domain adaptation (UDA) for person ReID. We evaluate the effectiveness of SNR on both DG-ReID and UDA-ReID by comparing with the state-of-the-art approaches in Table \ref{tab:STO}.    

\textbf{Domain generalizable person ReID} is very attractive in practical applications, which supports ``train once and run everywhere". However, there are very few works in this field \cite{song2019generalizable,jia2019frustratingly,zhou2019omni,kumar2019fairest}. Thanks to the exploration of the style normalization and restitution, our scheme \emph{SNR(Ours)} significantly outperforms the second best method \emph{OSNet-IBN} \cite{zhou2019omni} by \textbf{6.9\%} and \textbf{7.8\%} for Market1501$\rightarrow$Duke and Duke$\rightarrow$Market1501 in mAP, respectively. \emph{OSNet-IBN} adds Instance Normalization (IN) to the lower layers of their proposed OSNet following \cite{pan2018two}. However, this does not overcome the intrinsic shortcoming of IN and is not optimal.

Song \etal \cite{song2019generalizable} also explore domain generalizable person ReID and propose a Domain-Invariant Mapping Network (\emph{DIMN}) to learn the mapping between a person image and its identity classifier with a meta-learning pipeline. We follow \cite{song2019generalizable} and train \emph{SNR} on the same five datasets (M+D+C+CUHK02\cite{li2013locally}+CUHK-SYSU\cite{xiao2016end}). \emph{SNR} outperforms \emph{DIMN} by \textbf{14.6\%/6.6\%/1.2\%/11.5\%} in mAP and \textbf{12.9\%/10.9\%/1.7\%/13.9\%} in Rank-1 on the PRID/GRID/VIPeR/i-LIDS.

\textbf{Unsupervised domain adaptation for ReID} has been extensively studied where the unlabeled target data is also used for training. We follow the most commonly-used source$\rightarrow$target setting \cite{zhong2019invariance,liu2019adaptive,zhou2019omni,yu2019unsupervised,yang2019patch} for comparison. We take \emph{SNR} (see Figure \ref{fig:flowchart}(a)) as the backbone followed by a domain adaptation strategy MAR \cite{yu2019unsupervised} for domain adaptation, which we denote as \emph{SNR(Ours)+MAR} \cite{yu2019unsupervised}. For comparison, we take our strong \emph{Baseline} as the backbone followed by MAR, which we denote as \emph{Baseline+MAR}, to evaluate the effectiveness of the proposed SNR modules. We can see that \emph{SNR(Ours)+MAR} \cite{yu2019unsupervised} significantly outperforms the second-best UDA ReID method by \textbf{3.8\%}, \textbf{3.4\%} in mAP for Market1501+Duke(U)$\rightarrow$Duke and Duke+Market1501(U)$\rightarrow$Market1501, respectively. In addition, \emph{SNR(Ours)+MAR} outperforms \emph{Baseline+MAR} by \textbf{22.9\%}, \textbf{24.5\%} in mAP. Similar trends can be found for MSMT17+Duke(U)$\rightarrow$Duke and MSMT17+Market1501(U)$\rightarrow$Market1501. 


In general, as a plug-and-play module, SNR clearly enhances the generalization capability of ReID networks.

\subsection{Extension}
\label{subsec:extension}

\noindent\textbf{Performance on Other Backbone.}
We add SNR into the recently proposed lightweight ReID network OSNet \cite{zhou2019omni} and observe that by simply inserting SNR modules between the OS-Blocks, the new scheme \emph{OSNet-SNR} outperforms their model \emph{OSNet-IBN} by \textbf{5.0\%} and \textbf{5.5\%} in mAP for M$\rightarrow$D and D$\rightarrow$M, respectively (see \textbf{Supplementary}).

\noindent\textbf{RGB-Infrared Cross-Modality Person ReID.}
To further demonstrate the capability of SNR in handling images with large style variations, we conduct experiment on a more challenging RGB-Infrared cross-modality person ReID task on benchmark dataset SYSU-MM01 \cite{wu2017rgb}. Our scheme which integrates SNR to \emph{Baseline} outperforms \emph{Baseline} significantly by \textbf{8.4\%, 8.2\%, 11.0\%}, and \textbf{11.5\%} in mAP under 4 different  settings, and also achieves the state-of-the-art performance  (see \textbf{Supplementary} for more details).





\vspace{-2mm}
\section{Conclusion}
In this paper, we propose a generalizable person ReID framework to enable effective ReID. A Style Normalization and Restitution (SNR) module is introduced to exploit the merit of Instance Normalization (IN) that filters out the interference from style variations, and restitute the identity-relevant features that are discarded by IN. To efficiently disentangle the identity-relevant and -irrelevant features, we further design a dual causality loss constraint in SNR. Extensive experiments on several benchmarks/settings demonstrate the effectiveness of SNR. Our framework with SNR embedded achieves the best performance on both domain generalization and unsupervised domain adaptation ReID. Moreover, we have also verified SNR's effectiveness on RGB-Infrared ReID task, and on another backbone.

\section{Acknowledgments}
This work was supported in part by NSFC under Grant U1908209, 61632001 and the National Key Research and Development Program of China 2018AAA0101400.

{\small
\bibliographystyle{ieee_fullname}
\bibliography{bib_reid}

\begin{thebibliography}{10}\itemsep=-1pt

\bibitem{almazan2018re}
Jon Almazan, Bojana Gajic, Naila Murray, and Diane Larlus.
\newblock Re-id done right: towards good practices for person
  re-identification.
\newblock {\em arXiv preprint arXiv:1801.05339}, 2018.

\bibitem{ba2016layer}
Jimmy~Lei Ba, Jamie~Ryan Kiros, and Geoffrey~E Hinton.
\newblock Layer normalization.
\newblock {\em arXiv preprint arXiv:1607.06450}, 2016.

\bibitem{dai2018cross}
Pingyang Dai, Rongrong Ji, Haibin Wang, Qiong Wu, and Yuyu Huang.
\newblock Cross-modality person re-identification with generative adversarial
  training.
\newblock In {\em IJCAI}, pages 677--683, 2018.

\bibitem{dalal2005histograms}
Navneet Dalal and Bill Triggs.
\newblock Histograms of oriented gradients for human detection.
\newblock 2005.

\bibitem{deng2018image}
Weijian Deng, Liang Zheng, Qixiang Ye, Guoliang Kang, Yi Yang, and Jianbin
  Jiao.
\newblock Image-image domain adaptation with preserved self-similarity and
  domain-dissimilarity for person re-identification.
\newblock In {\em CVPR}, 2018.

\bibitem{dumoulin2016learned}
Vincent Dumoulin, Jonathon Shlens, and Manjunath Kudlur.
\newblock A learned representation for artistic style.
\newblock {\em ICLR}, 2017.

\bibitem{fan2018unsupervised}
Hehe Fan, Liang Zheng, Chenggang Yan, and Yi Yang.
\newblock Unsupervised person re-identification: Clustering and fine-tuning.
\newblock {\em ACM Transactions on Multimedia Computing, Communications, and
  Applications (TOMM)}, 2018.

\bibitem{Fu2018SelfsimilarityGA}
Yang Fu, Yunchao Wei, Guanshuo Wang, Xi Zhou, Honghui Shi, and Thomas~S. Huang.
\newblock Self-similarity grouping: A simple unsupervised cross domain
  adaptation approach for person re-identification.
\newblock {\em ICCV}, abs/1811.10144, 2019.

\bibitem{fu2019horizontal}
Yang Fu, Yunchao Wei, Yuqian Zhou, et~al.
\newblock Horizontal pyramid matching for person re-identification.
\newblock In {\em AAAI}, 2019.

\bibitem{ge2018fd}
Yixiao Ge, Zhuowan Li, Haiyu Zhao, et~al.
\newblock Fd-gan: Pose-guided feature distilling gan for robust person
  re-identification.
\newblock In {\em NeurIPS}, 2018.

\bibitem{gray2008viewpoint}
Douglas Gray and Hai Tao.
\newblock Viewpoint invariant pedestrian recognition with an ensemble of
  localized features.
\newblock In {\em ECCV}, pages 262--275. Springer, 2008.

\bibitem{hao2019hsme}
Yi Hao, Nannan Wang, Jie Li, and Xinbo Gao.
\newblock Hsme: Hypersphere manifold embedding for visible thermal person
  re-identification.
\newblock In {\em AAAI}, volume~33, pages 8385--8392, 2019.

\bibitem{he2016deep}
Kaiming He, Xiangyu Zhang, Shaoqing Ren, et~al.
\newblock Deep residual learning for image recognition.
\newblock In {\em CVPR}, 2016.

\bibitem{hermans2017defense}
Alexander Hermans, Lucas Beyer, and Bastian Leibe.
\newblock In defense of the triplet loss for person re-identification.
\newblock {\em arXiv preprint arXiv:1703.07737}, 2017.

\bibitem{hirzer2011person}
Martin Hirzer, Csaba Beleznai, Peter~M Roth, and Horst Bischof.
\newblock Person re-identification by descriptive and discriminative
  classification.
\newblock In {\em SCIA}, pages 91--102. Springer, 2011.

\bibitem{hu2018squeeze}
Jie Hu, Li Shen, and Gang Sun.
\newblock Squeeze-and-excitation networks.
\newblock In {\em CVPR}, pages 7132--7141, 2018.

\bibitem{huang2017arbitrary}
Xun Huang and Serge Belongie.
\newblock Arbitrary style transfer in real-time with adaptive instance
  normalization.
\newblock In {\em ICCV}, pages 1501--1510, 2017.

\bibitem{ioffe2015batch}
Sergey Ioffe and Christian Szegedy.
\newblock Batch normalization: Accelerating deep network training by reducing
  internal covariate shift.
\newblock {\em ICML}, 2015.

\bibitem{jia2019frustratingly}
Jieru Jia, Qiuqi Ruan, and Timothy~M Hospedales.
\newblock Frustratingly easy person re-identification: Generalizing person
  re-id in practice.
\newblock {\em BMVC}, 2019.

\bibitem{jin2020uncertainty}
Xin Jin, Cuiling Lan, Wenjun Zeng, and Zhibo Chen.
\newblock Uncertainty-aware multi-shot knowledge distillation for image-based
  object re-identification.
\newblock In {\em AAAI}, 2020.

\bibitem{jin2020semantics}
Xin Jin, Cuiling Lan, Wenjun Zeng, Guoqiang Wei, and Zhibo Chen.
\newblock Semantics-aligned representation learning for person
  re-identification.
\newblock In {\em AAAI}, 2020.

\bibitem{khosla2012undoing}
Aditya Khosla, Tinghui Zhou, Tomasz Malisiewicz, Alexei~A Efros, and Antonio
  Torralba.
\newblock Undoing the damage of dataset bias.
\newblock In {\em ECCV}, pages 158--171. Springer, 2012.

\bibitem{kingma2014adam}
Diederik~P Kingma and Jimmy Ba.
\newblock Adam: A method for stochastic optimization.
\newblock In {\em ICLR}, 2014.

\bibitem{kumar2019fairest}
Devinder Kumar, Parthipan Siva, Paul Marchwica, and Alexander Wong.
\newblock Fairest of them all: Establishing a strong baseline for cross-domain
  person reid.
\newblock {\em arXiv preprint arXiv:1907.12016}, 2019.

\bibitem{li2017learning}
Dangwei Li, Xiaotang Chen, Zhang Zhang, et~al.
\newblock Learning deep context-aware features over body and latent parts for
  person re-identification.
\newblock In {\em CVPR}, 2017.

\bibitem{li2018learning}
Da Li, Yongxin Yang, Yi-Zhe Song, and Timothy~M Hospedales.
\newblock Learning to generalize: Meta-learning for domain generalization.
\newblock In {\em AAAI}, 2018.

\bibitem{li2013locally}
Wei Li and Xiaogang Wang.
\newblock Locally aligned feature transforms across views.
\newblock In {\em CVPR}, pages 3594--3601, 2013.

\bibitem{li2014deepreid}
Wei Li, Rui Zhao, Lu Tian, et~al.
\newblock Deepreid: Deep filter pairing neural network for person
  re-identification.
\newblock In {\em CVPR}, 2014.

\bibitem{li2016revisiting}
Yanghao Li, Naiyan Wang, Jianping Shi, Jiaying Liu, and Xiaodi Hou.
\newblock Revisiting batch normalization for practical domain adaptation.
\newblock {\em arXiv preprint arXiv:1603.04779}, 2016.

\bibitem{li2018adaptation}
Yu-Jhe Li, Fu-En Yang, Yen-Cheng Liu, Yu-Ying Yeh, Xiaofei Du, and Yu-Chiang
  Frank~Wang.
\newblock Adaptation and re-identification network: An unsupervised deep
  transfer learning approach to person re-identification.
\newblock In {\em CVPR workshops}, 2018.

\bibitem{liao2015person}
Shengcai Liao, Yang Hu, Xiangyu Zhu, and Stan~Z Li.
\newblock Person re-identification by local maximal occurrence representation
  and metric learning.
\newblock In {\em CVPR}, 2015.

\bibitem{liao2015efficient}
Shengcai Liao and Stan~Z Li.
\newblock Efficient psd constrained asymmetric metric learning for person
  re-identification.
\newblock In {\em ICCV}, pages 3685--3693, 2015.

\bibitem{lin2016cross}
Liang Lin, Guangrun Wang, Wangmeng Zuo, Xiangchu Feng, and Lei Zhang.
\newblock Cross-domain visual matching via generalized similarity measure and
  feature learning.
\newblock {\em TPAMI}, 39(6):1089--1102, 2016.

\bibitem{lin2018multi}
Shan Lin, Haoliang Li, Chang-Tsun Li, and Alex~Chichung Kot.
\newblock Multi-task mid-level feature alignment network for unsupervised
  cross-dataset person re-identification.
\newblock {\em BMVC}, 2018.

\bibitem{liu2019adaptive}
Jiawei Liu, Zheng-Jun Zha, Di Chen, Richang Hong, and Meng Wang.
\newblock Adaptive transfer network for cross-domain person re-identification.
\newblock In {\em CVPR}, 2019.

\bibitem{loy2010time}
Chen~Change Loy, Tao Xiang, and Shaogang Gong.
\newblock Time-delayed correlation analysis for multi-camera activity
  understanding.
\newblock {\em IJCV}, 90(1):106--129, 2010.

\bibitem{luo2019bag}
Hao Luo, Youzhi Gu, Xingyu Liao, Shenqi Lai, and Wei Jiang.
\newblock Bag of tricks and a strong baseline for deep person
  re-identification.
\newblock In {\em CVPR workshops}, 2019.

\bibitem{luo2018differentiable}
Ping Luo, Jiamin Ren, Zhanglin Peng, Ruimao Zhang, and Jingyu Li.
\newblock Differentiable learning-to-normalize via switchable normalization.
\newblock {\em ICLR}, 2019.

\bibitem{maaten2008visualizing}
Laurens van~der Maaten and Geoffrey Hinton.
\newblock Visualizing data using t-sne.
\newblock {\em JMLR}, 2008.

\bibitem{muandet2013domain}
Krikamol Muandet, David Balduzzi, and Bernhard Sch{\"o}lkopf.
\newblock Domain generalization via invariant feature representation.
\newblock In {\em ICML}, pages 10--18, 2013.

\bibitem{pan2018two}
Xingang Pan, Ping Luo, Jianping Shi, and Xiaoou Tang.
\newblock Two at once: Enhancing learning and generalization capacities via
  ibn-net.
\newblock In {\em ECCV}, 2018.

\bibitem{qi2019novel}
Lei Qi, Lei Wang, Jing Huo, Luping Zhou, Yinghuan Shi, and Yang Gao.
\newblock A novel unsupervised camera-aware domain adaptation framework for
  person re-identification.
\newblock {\em ICCV}, 2019.

\bibitem{qian2018pose}
Xuelin Qian, Yanwei Fu, Wenxuan Wang, et~al.
\newblock Pose-normalized image generation for person re-identification.
\newblock In {\em ECCV}, 2018.

\bibitem{shankar2018generalizing}
Shiv Shankar, Vihari Piratla, Soumen Chakrabarti, Siddhartha Chaudhuri, Preethi
  Jyothi, and Sunita Sarawagi.
\newblock Generalizing across domains via cross-gradient training.
\newblock {\em ICLR}, 2018.

\bibitem{song2019generalizable}
Jifei Song, Yongxin Yang, Yi-Zhe Song, Tao Xiang, and Timothy~M Hospedales.
\newblock Generalizable person re-identification by domain-invariant mapping
  network.
\newblock In {\em CVPR}, 2019.

\bibitem{song2018unsupervised}
Liangchen Song, Cheng Wang, Lefei Zhang, Bo Du, Qian Zhang, Chang Huang, and
  Xinggang Wang.
\newblock Unsupervised domain adaptive re-identification: Theory and practice.
\newblock {\em arXiv preprint arXiv:1807.11334}, 2018.

\bibitem{su2017pose}
Chi Su, Jianing Li, Shiliang Zhang, et~al.
\newblock Pose-driven deep convolutional model for person re-identification.
\newblock In {\em ICCV}, 2017.

\bibitem{sun2018beyond}
Yifan Sun, Liang Zheng, Yi Yang, Qi Tian, and Shengjin Wang.
\newblock Beyond part models: Person retrieval with refined part pooling (and a
  strong convolutional baseline).
\newblock In {\em ECCV}, pages 480--496, 2018.

\bibitem{szegedy2016rethinking}
Christian Szegedy, Vincent Vanhoucke, Sergey Ioffe, Jon Shlens, and Zbigniew
  Wojna.
\newblock Rethinking the inception architecture for computer vision.
\newblock In {\em CVPR}, 2016.

\bibitem{tang2019unsupervised}
Haotian Tang, Yiru Zhao, and Hongtao Lu.
\newblock Unsupervised person re-identification with iterative self-supervised
  domain adaptation.
\newblock In {\em CVPR workshops}, 2019.

\bibitem{ulyanov2016instance}
Dmitry Ulyanov, Andrea Vedaldi, and Victor Lempitsky.
\newblock Instance normalization: The missing ingredient for fast stylization.
\newblock {\em arXiv preprint arXiv:1607.08022}, 2016.

\bibitem{ulyanov2017improved}
Dmitry Ulyanov, Andrea Vedaldi, and Victor Lempitsky.
\newblock Improved texture networks: Maximizing quality and diversity in
  feed-forward stylization and texture synthesis.
\newblock In {\em CVPR}, pages 6924--6932, 2017.

\bibitem{wang2018transferable}
Jingya Wang, Xiatian Zhu, Shaogang Gong, and Wei Li.
\newblock Transferable joint attribute-identity deep learning for unsupervised
  person re-identification.
\newblock In {\em CVPR}, 2018.

\bibitem{wang2018resource}
Yan Wang, Lequn Wang, Yurong You, Xu Zou, Vincent Chen, Serena Li, Gao Huang,
  Bharath Hariharan, and Kilian~Q Weinberger.
\newblock Resource aware person re-identification across multiple resolutions.
\newblock In {\em CVPR}, 2018.

\bibitem{wang2019learning}
Zhixiang Wang, Zheng Wang, Yinqiang Zheng, Yung-Yu Chuang, and Shin'ichi Satoh.
\newblock Learning to reduce dual-level discrepancy for infrared-visible person
  re-identification.
\newblock In {\em CVPR}, pages 618--626, 2019.

\bibitem{wei2018person}
Longhui Wei, Shiliang Zhang, Wen Gao, and Qi Tian.
\newblock Person transfer {GAN} to bridge domain gap for person
  re-identification.
\newblock In {\em CVPR}, 2018.

\bibitem{wei2009associating}
Zheng Wei-Shi, Gong Shaogang, and Xiang Tao.
\newblock Associating groups of people.
\newblock In {\em BMVC}, pages 23--1, 2009.

\bibitem{wu2017rgb}
Ancong Wu, Wei-Shi Zheng, Hong-Xing Yu, Shaogang Gong, and Jianhuang Lai.
\newblock Rgb-infrared cross-modality person re-identification.
\newblock In {\em ICCV}, pages 5380--5389, 2017.

\bibitem{xiao2016end}
Tong Xiao, Shuang Li, Bochao Wang, Liang Lin, and Xiaogang Wang.
\newblock End-to-end deep learning for person search.
\newblock {\em arXiv preprint arXiv:1604.01850}, 2:2, 2016.

\bibitem{yang2019patch}
Qize Yang, Hong-Xing Yu, Ancong Wu, and Wei-Shi Zheng.
\newblock Patch-based discriminative feature learning for unsupervised person
  re-identification.
\newblock In {\em CVPR}, 2019.

\bibitem{ye2018hierarchical}
Mang Ye, Xiangyuan Lan, Jiawei Li, and Pong~C Yuen.
\newblock Hierarchical discriminative learning for visible thermal person
  re-identification.
\newblock In {\em AAAI}, 2018.

\bibitem{ye2018visible}
Mang Ye, Zheng Wang, Xiangyuan Lan, and Pong~C Yuen.
\newblock Visible thermal person re-identification via dual-constrained
  top-ranking.
\newblock In {\em IJCAI}, pages 1092--1099, 2018.

\bibitem{yu2017cross}
Hong-Xing Yu, Ancong Wu, and Wei-Shi Zheng.
\newblock Cross-view asymmetric metric learning for unsupervised person
  re-identification.
\newblock In {\em ICCV}, 2017.

\bibitem{yu2019unsupervised}
Hong-Xing Yu, Wei-Shi Zheng, Ancong Wu, Xiaowei Guo, Shaogang Gong, and
  Jian-Huang Lai.
\newblock Unsupervised person re-identification by soft multilabel learning.
\newblock In {\em CVPR}, 2019.

\bibitem{zhang2016learning}
Li Zhang, Tao Xiang, and Shaogang Gong.
\newblock Learning a discriminative null space for person re-identification.
\newblock In {\em CVPR}, 2016.

\bibitem{zhang2019self}
Xinyu Zhang, Jiewei Cao, Chunhua Shen, and Mingyu You.
\newblock Self-training with progressive augmentation for unsupervised
  cross-domain person re-identification.
\newblock In {\em ICCV}, 2019.

\bibitem{zhang2019DSA}
Zhizheng Zhang, Cuiling Lan, Wenjun Zeng, et~al.
\newblock Densely semantically aligned person re-identification.
\newblock In {\em CVPR}, 2019.

\bibitem{zhao2017spindle}
Haiyu Zhao, Maoqing Tian, Shuyang Sun, et~al.
\newblock Spindle net: Person re-identification with human body region guided
  feature decomposition and fusion.
\newblock In {\em CVPR}, 2017.

\bibitem{zheng2015scalable}
Liang Zheng, Liyue Shen, et~al.
\newblock Scalable person re-identification: A benchmark.
\newblock In {\em ICCV}, 2015.

\bibitem{zheng2011person}
Wei-Shi Zheng, Shaogang Gong, and Tao Xiang.
\newblock Person re-identification by probabilistic relative distance
  comparison.
\newblock In {\em CVPR}, 2011.

\bibitem{zheng2017unlabeled}
Zhedong Zheng, Liang Zheng, and Yi Yang.
\newblock Unlabeled samples generated by gan improve the person
  re-identification baseline in vitro.
\newblock In {\em ICCV}, 2017.

\bibitem{zhong2018generalizing}
Zhun Zhong, Liang Zheng, Shaozi Li, and Yi Yang.
\newblock Generalizing a person retrieval model hetero-and homogeneously.
\newblock In {\em ECCV}, 2018.

\bibitem{zhong2019invariance}
Zhun Zhong, Liang Zheng, Zhiming Luo, Shaozi Li, and Yi Yang.
\newblock Invariance matters: Exemplar memory for domain adaptive person
  re-identification.
\newblock In {\em CVPR}, pages 598--607, 2019.

\bibitem{zhong2018camstyle}
Zhun Zhong, Liang Zheng, Zhedong Zheng, Shaozi Li, and Yi Yang.
\newblock Camstyle: A novel data augmentation method for person
  re-identification.
\newblock {\em TIP}, 2018.

\bibitem{zhou2019omni}
Kaiyang Zhou, Yongxin Yang, Andrea Cavallaro, et~al.
\newblock Omni-scale feature learning for person re-identification.
\newblock {\em ICCV}, 2019.

\end{thebibliography}
}

\clearpage

\appendix
  \renewcommand\thesection{\arabic{section}}

\noindent{\LARGE \textbf{Appendix}}
\vspace{5mm}

\section{Implementation Details}

\paragraph{Network Details.} We use ResNet-50 \cite{he2016deep,almazan2018re,zhang2019DSA,luo2019bag} as our base network for both baselines and our schemes. We build a strong baseline \emph{Baseline} with some commonly used tricks integrated. Similar to \cite{almazan2018re,zhang2019DSA,luo2019bag}, the last spatial down-sample operation in the last Conv block is removed. The proposed SNR module is added after the last layer of each convolutional block/stage of the first four stages. The input image resolution is 256$\times$128.

\paragraph{Data Augmentation.} We use the commonly used data augmentation strategies of random cropping \cite{wang2018resource,zhang2019DSA}, horizontal flipping, and label smoothing regularization \cite{szegedy2016rethinking}. To enhance the generalization ability, we further incorporate some useful data augmentation tricks, such as color jittering and disabling random erasing (REA) \cite{luo2019bag,zhou2019omni}. REA hurts models in cross-domain ReID task \cite{luo2019bag,kumar2019fairest}, because REA which masks the regions of training images makes the model learn more knowledge in the training source domain. It causes the model to perform worse in the unseen target domain.

\paragraph{Training Details for Domain Generalization.} Following \cite{hermans2017defense}, a batch is formed by first randomly sampling $P$ identities. For each identity, we sample $K$ images. Then the batch size is $B=P\times K$. We set $P=24$ and $K=4$ (\ieno, batch size $B=P\times K=96$. 

We use the Adam optimizer \cite{kingma2014adam} for model optimization. Similar to \cite{luo2019bag,zhang2019DSA}, we first warm up the model for 20 epochs with a linear growth learning rate from 8$\times$10$^{-6}$ to 8$\times$10$^{-4}$. Then we set the initial learning rate as 8$\times$10$^{-4}$ and optimize the Adam optimizer with a weight decay of 5$\times$10$^{-4}$. The learning rate is decayed by a factor of 0.5 for every 40 epochs. Our model (here we use ResNet-50 as our backbone) with SNR converges well after training of 280 epochs and we use it for evaluating the generalization performance on target datasets. All our models are implemented on PyTorch and trained on a single 32G NVIDIA-V100 GPU.

\paragraph{Training Details for Domain Adaptation.} For unsupervised domain adaptation person ReID, we combine our network with the unsupervised ReID approach MAR \cite{yu2019unsupervised} for fine-tuning on the unlabelled target domain data. MAR \cite{yu2019unsupervised} plays the role of assigning psudeo labels by hard negative mining, which facilitates the fine-tuning of base network. Similar to \cite{yu2019unsupervised}, during the fine-tuning, both source labeled data and target unlabelled data are jointly used for effective joint training. Specifically, during fine-tuning, a training batch of size 96 is composed of 1) labeled source data (size $B_{1} = P \times K=48$, where $P=12, K=4$), and 2) unlabeled target data (size $B_{2} = 48$). For the labeled source data, we optimize the network with the ReID loss $\mathcal{L}_{ReID}$ and the proposed dual causality loss $\mathcal{L}_{SNR}$. For the unlabeled target data, we follow the adaptation strategy of MAR \cite{yu2019unsupervised} to assign a pseudo soft multilabel for each sample and utilize these pseudo labels to perform soft multilabel-guided hard negative mining for training. We fine-tune the network also with the Adam optimizer \cite{kingma2014adam} with a initial learning rate of 1$\times$10$^{-5}$ for 200 epochs. We optimize the Adam optimizer with a weight decay of 5$\times$10$^{-4}$. The learning rate is decayed by a factor of 0.5 at 50, 100 and 150 epochs.

\noindent\textbf{Why do we perform disentanglement only on channel level?} We perform feature disentanglement only on channel level for \textbf{two reasons}: \textbf{1)} Those identity-irrelevant style factors (\egno, illumination, contrast, saturation) are typically regarded as spatially consistent, which are hard to disentangle by spatial-attention. \textbf{2)} In our SNR, ``disentanglement'' aims at better ``restitution'' of the lost discriminative information due to Instance Normalization (IN). IN reduces style discrepancy of input features by performing normalization across spatial dimensions independently for each channel, where the normalization parameters are the same across different spatial positions. To be consistent with IN, we disentangle the features and restitute the identity-relevant ones to the normalized features on channel level.



\section{Details of Datasets}
\begin{table}[htbp]
  \centering
  \scriptsize
  \caption{Details about the ReID datasets.}
    \begin{tabular}{c|c|c|c|c}
    \toprule
    Datasets & Identities & Images & Cameras & Scene \\
    \midrule
    Market1501 \cite{zheng2015scalable} & 1501  & 32668 & 6     & outdoor \\
    DukeMTMC-reID \cite{zheng2017unlabeled} & 1404  & 32948 & 8     & outdoor \\
    CUHK03 \cite{li2014deepreid} & 1467  & 28192 & 2     & indoor \\
    MSMT17 \cite{wei2018person} & 4101  & 126142 & 15    & outdoor, indoor \\
    VIPeR \cite{gray2008viewpoint} & 632   & 1264  & 2     & outdoor \\
    PRID2011 \cite{hirzer2011person} & 385   & 1134  & 2     & outdoor \\
    GRID \cite{loy2010time} & 250   & 500   & 2     & indoor \\
    i-LIDS \cite{wei2009associating} & 119   & 476   & N/A   & indoor \\
    \bottomrule
    \end{tabular}%
  \label{tab:datasets}%
\end{table}%

\begin{table*}[htbp]
  \centering
  \scriptsize
  \caption{Performance (\%) comparisons of our scheme and others to demonstrate the effectiveness of our SNR module for generalizable person ReID. The rows denote source dataset(s) for training and the columns correspond to different target datasets for testing. We mask the results of supervised ReID by gray where the testing domain has been seen in training. Note that we show the total number of source training images by data num..} 
  \vspace{-3mm}
    \begin{tabular}{c|l|cccccccccccc}
    \multicolumn{1}{r}{} & \multicolumn{1}{r}{} &       &       &       &       &       &       &       &       &       &       &       &  \\
    \midrule
    \multirow{2}[2]{*}{Source} & \multicolumn{1}{c|}{\multirow{2}[2]{*}{Method}} & \multicolumn{2}{c}{Target: Market1501} & \multicolumn{2}{c}{Target: Duke} & \multicolumn{2}{c}{Target: PRID} & \multicolumn{2}{c}{Target: GRID} & \multicolumn{2}{c}{Target: VIPeR} & \multicolumn{2}{c}{Target: iLIDs} \\
          &       & mAP   & Rank-1 & mAP   & Rank-1 & mAP   & Rank-1 & mAP   & Rank-1 & mAP   & Rank-1 & mAP   & Rank-1 \\
    \midrule
    \multirow{6}[2]{*}{\begin{tabular}[c]{@{}c@{}}Market1501 (M)\\ data num. ~32.6k\end{tabular}} & Baseline & \cellcolor[rgb]{ .906,  .902,  .902}82.8 & \cellcolor[rgb]{ .906,  .902,  .902}93.2 & 19.8  & 35.3  & 13.7  & 6.0     & 25.8  & 16.0    & 37.6  & 28.5  & 61.5  & 53.3 \\
          & Baseline-A-IN & \cellcolor[rgb]{ .906,  .902,  .902}75.3 & \cellcolor[rgb]{ .906,  .902,  .902}89.8 & 24.1  & 42.7  & 33.9  & 21.0    & 35.6  & 27.2  & \textcolor[rgb]{ .357,  .608,  .835}{38.1} & \textcolor[rgb]{ .357,  .608,  .835}{29.1} & \textcolor[rgb]{ .357,  .608,  .835}{64.2} & \textcolor[rgb]{ .357,  .608,  .835}{55.0} \\
          & Baseline-IBN & \cellcolor[rgb]{ .906,  .902,  .902}81.1 & \cellcolor[rgb]{ .906,  .902,  .902}92.2 & 21.5  & 39.2  & 19.1  & 12.0    & 27.5  & 19.2  & 32.1  & 23.4  & 58.3  & 48.3 \\
          & Baseline-A-SN & \cellcolor[rgb]{ .906,  .902,  .902}\textcolor[rgb]{ .357,  .608,  .835}{83.2} & \cellcolor[rgb]{ .906,  .902,  .902}\textcolor[rgb]{ .357,  .608,  .835}{93.9} & 20.1  & 38.0    & \textcolor[rgb]{ .357,  .608,  .835}{35.4} & \textcolor[rgb]{ .357,  .608,  .835}{25.0} & 29.0    & 22.0    & 32.2  & 23.4  & 53.4  & 43.3 \\
          & Baseline-IN & \cellcolor[rgb]{ .906,  .902,  .902}79.5 & \cellcolor[rgb]{ .906,  .902,  .902}90.9 & \textcolor[rgb]{ .357,  .608,  .835}{25.1} & \textcolor[rgb]{ .357,  .608,  .835}{44.9} & 35.0    & \textcolor[rgb]{ .357,  .608,  .835}{25.0} & \textcolor[rgb]{ .357,  .608,  .835}{35.7} & \textcolor[rgb]{ .357,  .608,  .835}{27.8} & 35.1  & 27.5  & 64.0    & 54.2 \\
          & \textbf{Baseline-SNR (Ours)}  & \cellcolor[rgb]{ .906,  .902,  .902}\textcolor[rgb]{ 1,  0,  0}{\textbf{84.7}} & \cellcolor[rgb]{ .906,  .902,  .902}\textcolor[rgb]{ 1,  0,  0}{\textbf{94.4}} & \textcolor[rgb]{ 1,  0,  0}{\textbf{33.6}} & \textcolor[rgb]{ 1,  0,  0}{\textbf{55.1}} & \textcolor[rgb]{ 1,  0,  0}{\textbf{42.2}} & \textcolor[rgb]{ 1,  0,  0}{\textbf{30.0}} & \textcolor[rgb]{ 1,  0,  0}{\textbf{36.7}} & \textcolor[rgb]{ 1,  0,  0}{\textbf{29.0}} & \textcolor[rgb]{ 1,  0,  0}{\textbf{42.3}} & \textcolor[rgb]{ 1,  0,  0}{\textbf{32.3}} & \textcolor[rgb]{ 1,  0,  0}{\textbf{65.6}} & \textcolor[rgb]{ 1,  0,  0}{\textbf{56.7}} \\
    \midrule
    \multirow{6}[2]{*}{\begin{tabular}[c]{@{}c@{}}Duke (D)\\ data num. ~32.9k\end{tabular}} & Baseline & 21.8  & 48.3  & \cellcolor[rgb]{ .906,  .902,  .902}71.2 & \cellcolor[rgb]{ .906,  .902,  .902}83.4 & 15.7  & 11.0    & 14.5  & 8.8   & \textcolor[rgb]{ .357,  .608,  .835}{37.0} & 26.9  & 68.3  & 58.3 \\
          & Baseline-A-IN & 26.5  & 56.0    & \cellcolor[rgb]{ .906,  .902,  .902}64.5 & \cellcolor[rgb]{ .906,  .902,  .902}78.9 & 38.6  & 29.0    & 19.6  & \textcolor[rgb]{ .357,  .608,  .835}{13.6} & 35.1  & \textcolor[rgb]{ .357,  .608,  .835}{27.2} & 67.4  & 56.7 \\
          & Baseline-IBN & 24.6  & 52.5  & \cellcolor[rgb]{ .906,  .902,  .902}69.5 & \cellcolor[rgb]{ .906,  .902,  .902}81.4 & 27.4  & 19.0    & 19.9  & 12.0    & 32.8  & 23.4  & 63.5  & 61.7 \\
          & Baseline-A-SN & 25.3  & 55.0    & \cellcolor[rgb]{ .906,  .902,  .902}\textcolor[rgb]{ 1,  0,  0}{\textbf{73.0}} & \cellcolor[rgb]{ .906,  .902,  .902}\textcolor[rgb]{1,  0,  0}{\textbf{85.9}} & \textcolor[rgb]{ .357,  .608,  .835}{41.4} & \textcolor[rgb]{ .357,  .608,  .835}{32.0} & 18.8  & 12.8  & 31.3  & 24.1  & 64.8  & 63.3 \\
          & Baseline-IN & \textcolor[rgb]{ .357,  .608,  .835}{27.2} & \textcolor[rgb]{ .357,  .608,  .835}{58.5} & \cellcolor[rgb]{ .906,  .902,  .902}68.9 & \cellcolor[rgb]{ .906,  .902,  .902}80.4 & 40.5  & 27.0    & \textcolor[rgb]{ .357,  .608,  .835}{20.3} & 13.2  & 34.6  & 26.3  & \textcolor[rgb]{ .357,  .608,  .835}{70.6} & \textcolor[rgb]{ .357,  .608,  .835}{65} \\
          & \textbf{Baseline-SNR (Ours)}  & \textcolor[rgb]{ 1,  0,  0}{\textbf{33.9}} & \textcolor[rgb]{ 1,  0,  0}{\textbf{66.7}} & \cellcolor[rgb]{ .906,  .902,  .902}\textcolor[rgb]{ .357,  .608,  .835}{{72.9}} & \cellcolor[rgb]{ .906,  .902,  .902}\textcolor[rgb]{ .357,  .608,  .835}{{84.4}} & \textcolor[rgb]{ 1,  0,  0}{\textbf{45.4}} & \textcolor[rgb]{ 1,  0,  0}{\textbf{35.0}} & \textcolor[rgb]{ 1,  0,  0}{\textbf{35.3}} & \textcolor[rgb]{ 1,  0,  0}{\textbf{26.0}} & \textcolor[rgb]{ 1,  0,  0}{\textbf{41.2}} & \textcolor[rgb]{ 1,  0,  0}{\textbf{32.6}} & \textcolor[rgb]{ 1,  0,  0}{\textbf{79.3}} & \textcolor[rgb]{ 1,  0,  0}{\textbf{68.7}} \\
    \midrule
    \multirow{6}[2]{*}{\begin{tabular}[c]{@{}c@{}}Market1501 (M)\\ + Duke (D) \\ data num. ~65.5k\end{tabular}} & Baseline & \cellcolor[rgb]{ .906,  .902,  .902}72.6 & \cellcolor[rgb]{ .906,  .902,  .902}88.2 & \cellcolor[rgb]{ .906,  .902,  .902}60.0 & \cellcolor[rgb]{ .906,  .902,  .902}77.8 & 14.8  & 9.0     & 23.1  & 15.2  & 39.4  & 30.4  & \textcolor[rgb]{ .357,  .608,  .835}{74.3} & 65.0 \\
          & Baseline-A-IN & \cellcolor[rgb]{ .906,  .902,  .902}76.5 & \cellcolor[rgb]{ .906,  .902,  .902}91.4 & \cellcolor[rgb]{ .906,  .902,  .902}62.2 & \cellcolor[rgb]{ .906,  .902,  .902}80.1 & 45.0    & 30.0    & 36.7  & 28.0    & 37.3  & 28.2  & 73.6  & \textcolor[rgb]{ .357,  .608,  .835}{65.2} \\
          & Baseline-IBN & \cellcolor[rgb]{ .906,  .902,  .902}74.6 & \cellcolor[rgb]{ .906,  .902,  .902}90.4 & \cellcolor[rgb]{ .906,  .902,  .902}62.3 & \cellcolor[rgb]{ .906,  .902,  .902}80.1 & 43.7  & 32.0    & 32.6  & 24.0    & 42.8  & 33.2  & 73.8  & 65.0 \\
          & Baseline-A-SN & \cellcolor[rgb]{ .906,  .902,  .902}73.1 & \cellcolor[rgb]{ .906,  .902,  .902}89.8 & \cellcolor[rgb]{ .906,  .902,  .902}61.7 & \cellcolor[rgb]{ .906,  .902,  .902}79.0 & 47.9  & \textcolor[rgb]{ .357,  .608,  .835}{37.0} & 28.0    & 21.6  & 38.0    & 28.8  & 68.1  & 61.7 \\
          & Baseline-IN & \cellcolor[rgb]{ .906,  .902,  .902}\textcolor[rgb]{ .357,  .608,  .835}{77.5} & \cellcolor[rgb]{ .906,  .902,  .902}\textcolor[rgb]{ .357,  .608,  .835}{91.6} & \cellcolor[rgb]{ .906,  .902,  .902}\textcolor[rgb]{ .357,  .608,  .835}{63.9} & \cellcolor[rgb]{ .906,  .902,  .902}\textcolor[rgb]{ .357,  .608,  .835}{81.5} & \textcolor[rgb]{ .357,  .608,  .835}{48.1} & 36.0    & \textcolor[rgb]{ .357,  .608,  .835}{39.2} & \textcolor[rgb]{ .357,  .608,  .835}{31.2} & \textcolor[rgb]{ .357,  .608,  .835}{43.8} & \textcolor[rgb]{ .357,  .608,  .835}{33.9} & 73.2  & 64.3 \\
          & \textbf{Baseline-SNR (Ours)}  & \cellcolor[rgb]{ .906,  .902,  .902}\textcolor[rgb]{ 1,  0,  0}{\textbf{80.3}} & \cellcolor[rgb]{ .906,  .902,  .902}\textcolor[rgb]{ 1,  0,  0}{\textbf{92.9}} & \cellcolor[rgb]{ .906,  .902,  .902}\textcolor[rgb]{ 1,  0,  0}{\textbf{67.2}} & \cellcolor[rgb]{ .906,  .902,  .902}\textcolor[rgb]{ 1,  0,  0}{\textbf{83.1}} & \textcolor[rgb]{ 1,  0,  0}{\textbf{57.9}} & \textcolor[rgb]{ 1,  0,  0}{\textbf{50.0}} & \textcolor[rgb]{ 1,  0,  0}{\textbf{41.3}} & \textcolor[rgb]{ 1,  0,  0}{\textbf{34.4}} & \textcolor[rgb]{ 1,  0,  0}{\textbf{46.7}} & \textcolor[rgb]{ 1,  0,  0}{\textbf{37.7}} & \textcolor[rgb]{ 1,  0,  0}{\textbf{85.2}} & \textcolor[rgb]{ 1,  0,  0}{\textbf{80.0}} \\
    \midrule
    \multirow{6}[2]{*}{\begin{tabular}[c]{@{}c@{}}Market1501 (M)\\ + Duke (D)\\ + CUHK03 (C)\\ data num. ~93.7k\end{tabular}} & Baseline & \cellcolor[rgb]{ .906,  .902,  .902}76.4 & \cellcolor[rgb]{ .906,  .902,  .902}89.8 & \cellcolor[rgb]{ .906,  .902,  .902}63.6 & \cellcolor[rgb]{ .906,  .902,  .902}79.0 & 27.0    & 19.0    & 25.7  & 18.4  & 46.3  & 36.4  & 77.1  & 66.3 \\
          & Baseline-A-IN & \cellcolor[rgb]{ .906,  .902,  .902}76.8 & \cellcolor[rgb]{ .906,  .902,  .902}90.7 & \cellcolor[rgb]{ .906,  .902,  .902}63.0 & \cellcolor[rgb]{ .906,  .902,  .902}81.3 & 55.6  & 44.0    & 40.8  & \textcolor[rgb]{ .357,  .608,  .835}{33.6} & \textcolor[rgb]{ .357,  .608,  .835}{50.9} & \textcolor[rgb]{ .357,  .608,  .835}{41.8} & 77.7  & 70.0 \\
          & Baseline-IBN & \cellcolor[rgb]{ .906,  .902,  .902}76.2 & \cellcolor[rgb]{ .906,  .902,  .902}\textcolor[rgb]{ .357,  .608,  .835}{91.3} & \cellcolor[rgb]{ .906,  .902,  .902}62.8 & \cellcolor[rgb]{ .906,  .902,  .902}80.5 & \textcolor[rgb]{ .357,  .608,  .835}{56.6} & \textcolor[rgb]{ .357,  .608,  .835}{48.0} & 40.9  & 31.2  & 48.4  & 38.9  & 76.9  & 68.3 \\
          & Baseline-A-SN & \cellcolor[rgb]{ .906,  .902,  .902}71.1 & \cellcolor[rgb]{ .906,  .902,  .902}89.3 & \cellcolor[rgb]{ .906,  .902,  .902}62.0 & \cellcolor[rgb]{ .906,  .902,  .902}78.8 & 55.4  & 46.0    & 34.1  & 26.4  & 50.3  & 39.8  & 79.6  & 71.7 \\
          & Baseline-IN & \cellcolor[rgb]{ .906,  .902,  .902}\textcolor[rgb]{ .357,  .608,  .835}{77.8} & \cellcolor[rgb]{ .906,  .902,  .902}\textcolor[rgb]{ .357,  .608,  .835}{91.3} & \cellcolor[rgb]{ .906,  .902,  .902}\textcolor[rgb]{ .357,  .608,  .835}{64.4} & \cellcolor[rgb]{ .906,  .902,  .902}\textcolor[rgb]{ .357,  .608,  .835}{81.6} & 56.4  & 47.0    & \textcolor[rgb]{ .357,  .608,  .835}{41.0} & 31.8  & 49.3  & 39.9  & \textcolor[rgb]{ .357,  .608,  .835}{80.9} & \textcolor[rgb]{ .357,  .608,  .835}{74.7} \\
          & \textbf{Baseline-SNR (Ours)}  & \cellcolor[rgb]{ .906,  .902,  .902}\textcolor[rgb]{ 1,  0,  0}{\textbf{81.2}} & \cellcolor[rgb]{ .906,  .902,  .902}\textcolor[rgb]{ 1,  0,  0}{\textbf{93.3}} & \cellcolor[rgb]{ .906,  .902,  .902}\textcolor[rgb]{ 1,  0,  0}{\textbf{68.4}} & \cellcolor[rgb]{ .906,  .902,  .902}\textcolor[rgb]{ 1,  0,  0}{\textbf{84.2}} & \textcolor[rgb]{ 1,  0,  0}{\textbf{60.9}} & \textcolor[rgb]{ 1,  0,  0}{\textbf{52.0}} & \textcolor[rgb]{ 1,  0,  0}{\textbf{45.2}} & \textcolor[rgb]{ 1,  0,  0}{\textbf{36.8}} & \textcolor[rgb]{ 1,  0,  0}{\textbf{52.3}} & \textcolor[rgb]{ 1,  0,  0}{\textbf{42.4}} & \textcolor[rgb]{ 1,  0,  0}{\textbf{91.0}} & \textcolor[rgb]{ 1,  0,  0}{\textbf{86.7}} \\
          
    \midrule
    \midrule
    
        \multirow{2}[0]{*}{\begin{tabular}[c]{@{}c@{}}MSMT17 (MT) \\ data num. ~126k\end{tabular}} & Baseline & {23.1} &{48.2} &{29.2}  & {47.6} & {16.4} &{11.0}    &{9.8}    & {5.6}  & {40.8}  & {30.1}  & {74.0}  &{66.7} \\

      & \textbf{Baseline-SNR (Ours)} & \textcolor[rgb]{ 1,  0,  0}{\textbf{40.9}} & \textcolor[rgb]{ 1,  0,  0}{\textbf{69.5}} & \textcolor[rgb]{ 1,  0,  0}{\textbf{49.9}} & \textcolor[rgb]{ 1,  0,  0}{\textbf{69.2}} & \cellcolor[rgb]{ 1,  1,  1}\textcolor[rgb]{ 1,  0,  0}{\textbf{48.4}} & \cellcolor[rgb]{ 1,  1,  1}\textcolor[rgb]{ 1,  0,  0}{\textbf{39.0}} & \cellcolor[rgb]{ 1,  1,  1}\textcolor[rgb]{ 1,  0,  0}{\textbf{30.3}} & \cellcolor[rgb]{ 1,  1,  1}\textcolor[rgb]{ 1,  0,  0}{\textbf{24.0}} & \cellcolor[rgb]{ 1,  1,  1}\textcolor[rgb]{ 1,  0,  0}{\textbf{57.2}} & \cellcolor[rgb]{ 1,  1,  1}\textcolor[rgb]{ 1,  0,  0}{\textbf{47.5}} & \cellcolor[rgb]{ 1,  1,  1}\textcolor[rgb]{ 1,  0,  0}{\textbf{87.7}} & \cellcolor[rgb]{ 1,  1,  1}\textcolor[rgb]{ 1,  0,  0}{\textbf{81.9}} \\
    
    \midrule
    
    \multirow{2}[0]{*}{\begin{tabular}[c]{@{}c@{}}M + D + C + MT\\ data num. ~220k\end{tabular}} & Baseline & \cellcolor[rgb]{ .906,  .902,  .902}{72.4} &\cellcolor[rgb]{ .906,  .902,  .902}{88.7} & \cellcolor[rgb]{ .906,  .902,  .902}{70.1}  & \cellcolor[rgb]{ .906,  .902,  .902}{83.8} & {39.0} &{28.0}    &{29.6}    & {20.8}  & {52.1}  & {41.5}  & {89.0}  &{85.0} \\
	
	& \textbf{Baseline-SNR (Ours)} &\cellcolor[rgb]{ .906,  .902,  .902} \textcolor[rgb]{ 1,  0,  0}{\textbf{82.3}} & \cellcolor[rgb]{ .906,  .902,  .902} \textcolor[rgb]{ 1,  0,  0}{\textbf{93.4}} & \cellcolor[rgb]{ .906,  .902,  .902} \textcolor[rgb]{ 1,  0,  0}{\textbf{73.2}} & \cellcolor[rgb]{ .906,  .902,  .902} \textcolor[rgb]{ 1,  0,  0}{\textbf{85.5}} & \cellcolor[rgb]{ 1,  1,  1}\textcolor[rgb]{ 1,  0,  0}{\textbf{60.0}} & \cellcolor[rgb]{ 1,  1,  1}\textcolor[rgb]{ 1,  0,  0}{\textbf{49.0}} & \cellcolor[rgb]{ 1,  1,  1}\textcolor[rgb]{ 1,  0,  0}{\textbf{41.3}} & \cellcolor[rgb]{ 1,  1,  1}\textcolor[rgb]{ 1,  0,  0}{\textbf{30.4}} & \cellcolor[rgb]{ 1,  1,  1}\textcolor[rgb]{ 1,  0,  0}{\textbf{65.0}} & \cellcolor[rgb]{ 1,  1,  1}\textcolor[rgb]{ 1,  0,  0}{\textbf{55.1}} & \cellcolor[rgb]{ 1,  1,  1}\textcolor[rgb]{ 1,  0,  0}{\textbf{91.9}} & \cellcolor[rgb]{ 1,  1,  1}\textcolor[rgb]{ 1,  0,  0}{\textbf{87.0}} \\

    \bottomrule
    \end{tabular}%
  \label{tab:ablationstudy}%
\end{table*}%

\begin{figure}
  \centerline{\includegraphics[width=1.0\linewidth]{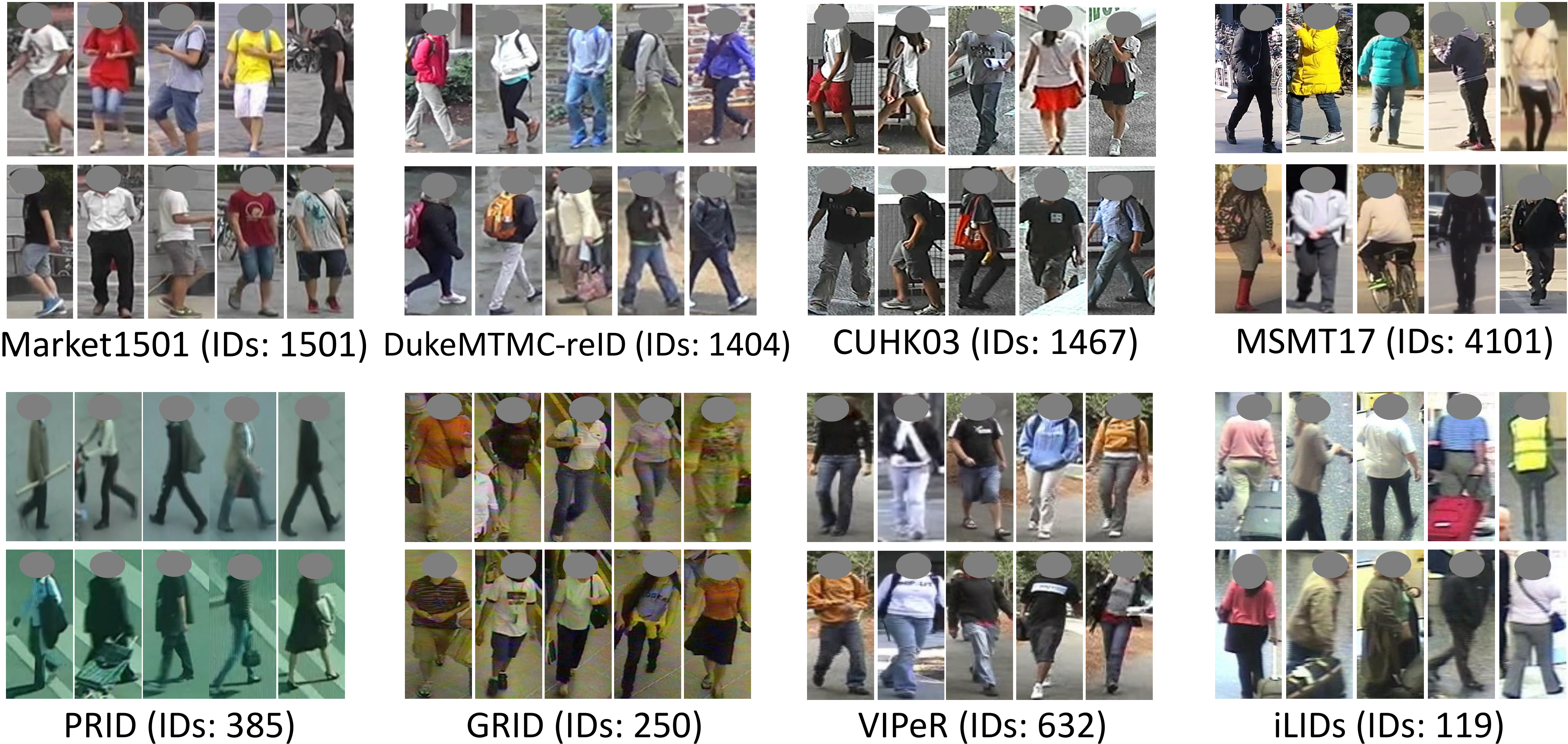}}
  \caption{Person images from different ReID datasets: Market-1501 \cite{zheng2015scalable}, DukeMTMC-reID \cite{zheng2017unlabeled}, CUHK03 \cite{li2014deepreid}, MSMT17 \cite{wei2018person}, and the four small-scale ReID datasets of PRID \cite{hirzer2011person}, GRID \cite{loy2010time}, VIPeR \cite{gray2008viewpoint}, and i-LIDS \cite{wei2009associating}. All images have been re-sized to 256$\times$128 for easier comparison. We observe there are obvious domain gaps/style discrepancies across different datasets, especially for PRID \cite{hirzer2011person} and GRID \cite{loy2010time}.}
\vspace{-3mm}
\label{fig:datasets}
\end{figure}

In Table \ref{tab:datasets}, we present the detailed information about the related person ReID datasets. Market1501 \cite{zheng2015scalable}, DukeMTMC-reID \cite{zheng2017unlabeled}, CUHK03 \cite{li2014deepreid}, and large-scale MSMT17 \cite{wei2018person} are the most commonly used datasets for fully supervised ReID \cite{zhang2019DSA,zhou2019omni} and unsupervised domain adaption ReID \cite{yu2019unsupervised,zhang2019self,Fu2018SelfsimilarityGA}. VIPeR \cite{gray2008viewpoint}, PRID2011 \cite{hirzer2011person}, GRID \cite{loy2010time}, and i-LIDS \cite{wei2009associating} are small ReID datasets
which could be used for evaluating cross-domain/generalizable
person ReID \cite{song2019generalizable,jia2019frustratingly,kumar2019fairest}. Market1501 \cite{zheng2015scalable} and DukeMTMC-reID \cite{zheng2017unlabeled} have pre-established test probe and test gallery splits which we use for our training and cross-test (\ieno, M$\rightarrow$D, D$\rightarrow$M). For the smaller datasets (VIPeR, PRID2011, GRID, and i-LIDS), we use the standard 10 random splits as in \cite{jia2019frustratingly,kumar2019fairest} for testing (the four small datasets are not involved in training). CUHK03 \cite{li2014deepreid} and MSMT17 \cite{wei2018person} are used for training.



We randomly pick up 10 identities from each ReID dataset and show them in Figure \ref{fig:datasets}. We observe that: 1) there is style discrepancy across datasets, which is rather obvious for PRID and GRID; 2) MSMT17 has large style variants within the same dataset. 


\begin{figure*}
  \centerline{\includegraphics[width=1.0\linewidth]{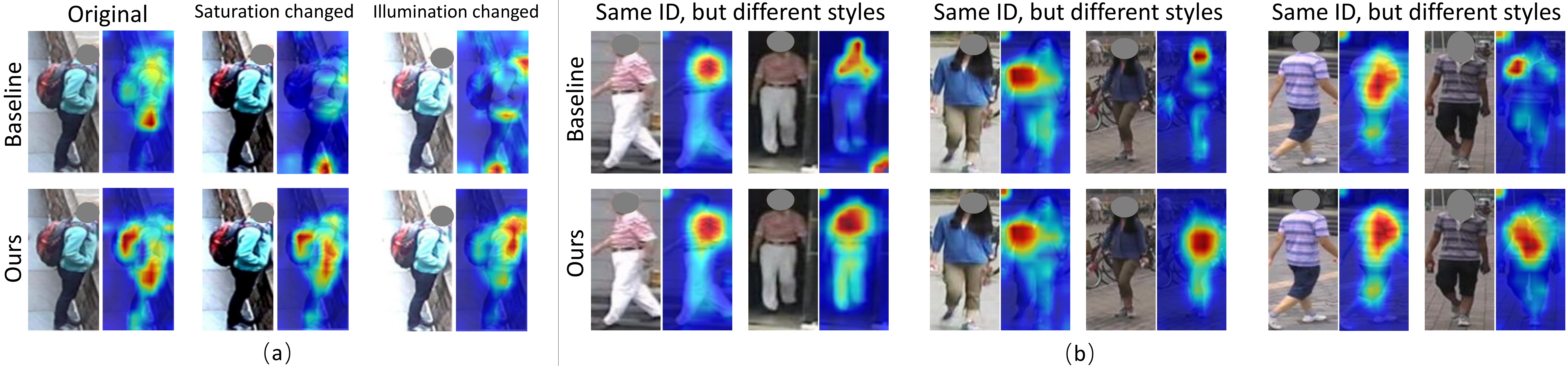}}
  \vspace{-2mm}
  \caption{Activation maps of our scheme (bottom) and the strong baseline  \emph{Baseline} (top) corresponding to images of varied styles. The maps of our method are more consistent/invariant to style variants.}
\label{fig:vis_ftp}
\vspace{-1mm}
\end{figure*}

\section{More Ablation Study Results}

We show more comparisons of our scheme and others to demonstrate the effectiveness of our SNR module for generalizable person ReID in Table \ref{tab:ablationstudy}.

We have observations consistent with those in our paper. 1) IN-related baselines bring generalization ability improvement but decrease the performance for the same-domain. 2) Our \emph{Baseline-SNR} achieves superior generalization capability thanks to the restitution of identity-relevant information by the SNR modules. 3) The generalization performance on unseen target domain increases consistently as the number of source datasets increases.

In Table \ref{tab:ablationstudy}, we also present the total number of source training images as marked by \emph{data num. N}. For the single source dataset settings, MSMT17 is the largest dataset, which contains 126k images while Market1501 or Duke has about 33K images. For the target testing datasets VIPeR and iLIDs, the performance of \emph{Baseline} trained by this large scale dataset MSMT17 is 3.8\% to 12.5\% higher than those trained by Market1501 or Duke in mAP. Generally, the increase of training data could improve the performance. However, the performance of \emph{Baseline} trained by MSMT17 has a rather low mAP accuracy of 9.8\% on the target dataset GRID, being even poorer than that trained on Market1501 (25.8\%) or Duke (14.5\%). For the target dataset PRID, similarly, MSMT17 does not provide clear superiority. These indicate that it is not always true that a larger amount of training data results in better performance. The domain gap between MSMT17 and GRID is larger than that between Market1510/Duke and GRID. To validate this, we analyze the feature divergence (FD, detailed descriptions can be found in Section \ref{sec:vis} below) between GRID and MSMT17, Market1501, Duke, respectively. We find that the divergence (here we calculate the feature divergence of the third convolutional block/stage within our \emph{Baseline-SNR} trained by combining all the four datasets) of Market1501 vs. GRID, Duke vs. GRID, MSMT17 vs. GRID are 2.17, 3.49, and 4.51, respectively. Note that the larger the FD value, the larger the feature discrepancy between the two domains. The domain gap between MSMT17 and GRID is larger than that between Market1501 (or Duke) and GRID. For the similar reason, we find that additionally adding MSMT17 as the source training data does not bring further performance improvement on GRID and PRID target datasets in our scheme \emph{Baseline-SNR} in comparison with the model trained by M+D+C source datasets.

\section{More Visualization Analysis}\label{sec:vis}

\noindent\textbf{More Feature Map Visualization.} 
In our paper, we compare the activation maps $\widetilde{F}^+$ of our scheme and those of the strong baseline scheme \emph{Baseline} by varying the styles of input images (\egno, contrast, illumination, saturation). Here, Figure \ref{fig:vis_ftp}(a) shows more visualization and Figure \ref{fig:vis_ftp}(b) shows visualization results on real images. We have the similar observations that the activation maps of our scheme are more consistent/invariant to style variants.




\begin{figure}
  \centerline{\includegraphics[width=0.96\linewidth]{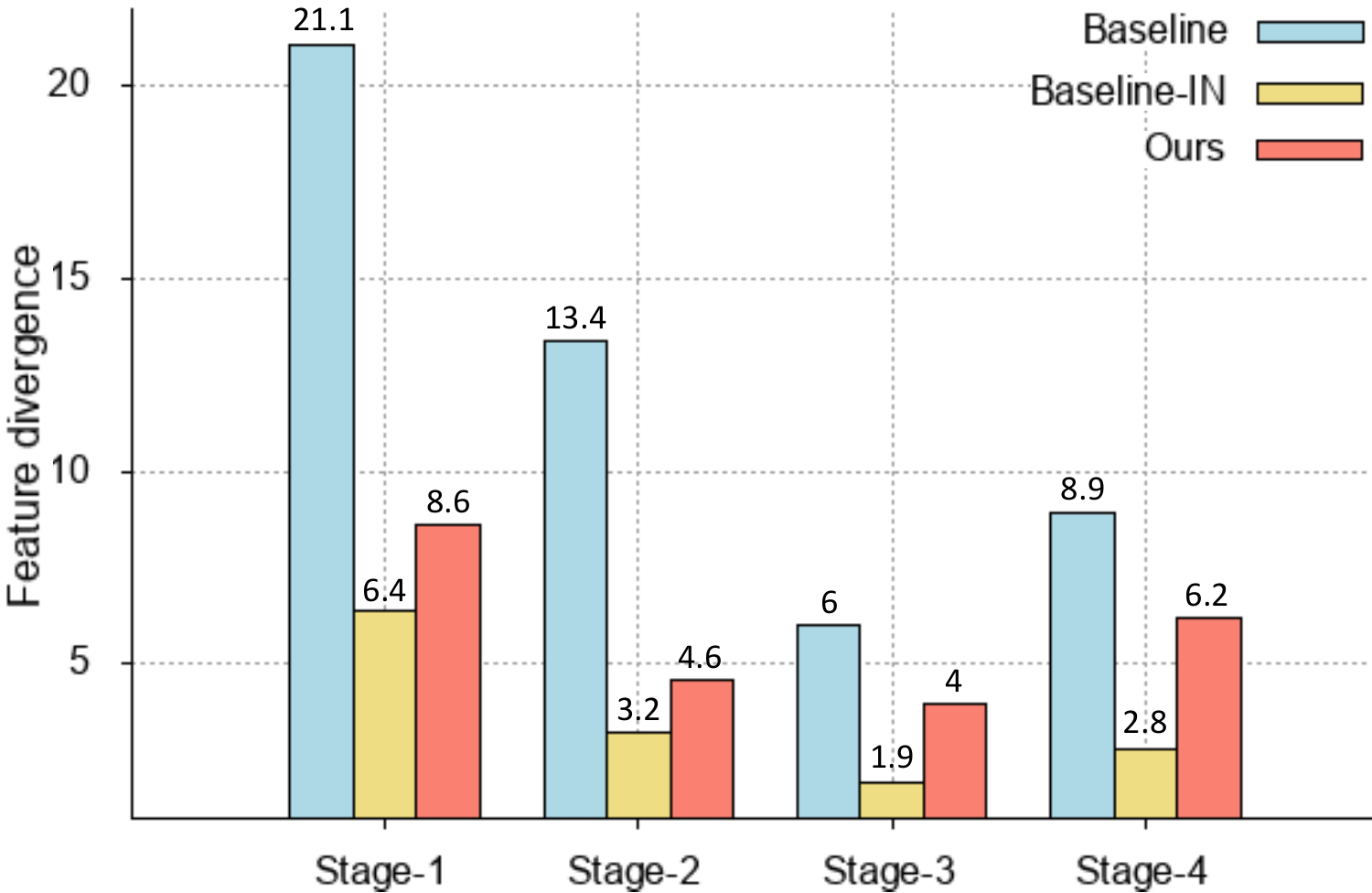}}
  \caption{Analysis of the feature divergence between two different domains, Market1501 and Duke.}
\label{fig:vis_fd}
\vspace{-3mm}
\end{figure}


\noindent\textbf{Feature Divergence Analysis.} We analyze the feature divergence between two datasets on three schemes: \emph{Baseline}, \emph{Baseline-IN}, and ours \emph{SNR}, respectively. Following \cite{pan2018two,li2016revisiting}, we use the symmetric KL divergence of features between domain A and B as the metric to measure feature divergence of the two domains. We train the models using Market1501 training dataset and evaluate the feature divergences between the test set of Market1501 and Duke (500 samples are randomly selected from each set). We calculate the feature divergence of the four convolutional blocks/stages respectively and show the results in Figure \ref{fig:vis_fd}.

We observe that the feature divergence (FD) is large for \emph{Baseline}. The introduction of IN as in scheme \emph{Baseline-IN} significantly reduces the FD on all the four stages. The FD of Stage-4 is higher than that in Stage-3. That is likely because Stage-4 is more related to high-level discriminative semantic features for distinguishing different identities. The discrimination may increase the feature divergence. Due to the introduction of the SNR modules, the FD on all convolutional blocks/stages is also significantly reduced in our scheme in comparison with \emph{Baseline}. It is higher than that of the scheme \emph{Baseline-IN} which is probably because the restitution of some identity-relevant features increases the discrimination capability and thus increases the FD. 
\begin{table*}[t] 
  \centering
  \scriptsize
  \caption{Performance (\%) comparisons with the state-of-the-art approaches for the Domain Generalizable Person ReID (top rows) and Unsupervised Domain Adaptation for Person ReID (bottom rows), respectively. ``(U)" denotes ``unlabeled". We mask the schemes of our \emph{Baseline} and our \emph{Baseline} with SNR modules (\ieno, \emph{SNR(Ours)}) by gray, with fair comparison between each pair to validate the effectiveness of SNR modules.}
    \begin{tabular}{c|cccccccc}
    \toprule
    \multicolumn{2}{c}{\multirow{2}[3]{*}{Method}} & \multirow{2}[3]{*}{Venue} & \multirow{2}[3]{*}{Source} & \multicolumn{2}{c}{Target: Duke} & \multirow{2}[3]{*}{Source} & \multicolumn{2}{c}{Taeget: Market1501} \\
\cmidrule{5-6}\cmidrule{8-9}    \multicolumn{2}{c}{} &       &       & mAP   & Rank-1    &       & mAP   & Rank-1 \\
    \midrule
    \midrule
    \multicolumn{1}{c|}{\multirow{10}[3]{*}{\begin{tabular}[c]{@{}c@{}} Domain\\Generalization \\(w/o using  \\ target data)\end{tabular}}}
    
          & OSNet-IBN \cite{zhou2019omni} & ICCV'19 & Market1501 & \textcolor[rgb]{ .357,  .608,  .835}{26.7}  & \textcolor[rgb]{ .357,  .608,  .835}{48.5}  & Duke  & \textcolor[rgb]{ .357,  .608,  .835}{26.1}  & \textcolor[rgb]{ .357,  .608,  .835}{57.7} \\
          & \cellcolor[rgb]{ .906,  .902,  .902}Baseline & \cellcolor[rgb]{ .906,  .902,  .902}This work & \cellcolor[rgb]{ .906,  .902,  .902}Market1501 & \cellcolor[rgb]{ .906,  .902,  .902}19.8 & \cellcolor[rgb]{ .906,  .902,  .902}35.3 & \cellcolor[rgb]{ .906,  .902,  .902}Duke & \cellcolor[rgb]{ .906,  .902,  .902}21.8 & \cellcolor[rgb]{ .906,  .902,  .902}48.3 \\

        & \cellcolor[rgb]{ .906,  .902,  .902}Baseline-IBN \cite{jia2019frustratingly} & \cellcolor[rgb]{ .906,  .902,  .902}BMVC'19 & \cellcolor[rgb]{ .906,  .902,  .902}Market1501 & \cellcolor[rgb]{ .906,  .902,  .902}21.5 & \cellcolor[rgb]{ .906,  .902,  .902}39.2 & \cellcolor[rgb]{ .906,  .902,  .902}Duke & \cellcolor[rgb]{ .906,  .902,  .902}24.6 & \cellcolor[rgb]{ .906,  .902,  .902}52.5 \\
          
          & \cellcolor[rgb]{ .906,  .902,  .902}\textbf{SNR(Ours)} & \cellcolor[rgb]{ .906,  .902,  .902}This work & \cellcolor[rgb]{ .906,  .902,  .902}Market1501 & \cellcolor[rgb]{ .906,  .902,  .902}\textcolor[rgb]{ 1,  0,  0}{\textbf{33.6}} & \cellcolor[rgb]{ .906,  .902,  .902}\textcolor[rgb]{ 1,  0,  0}{\textbf{55.1}} & \cellcolor[rgb]{ .906,  .902,  .902}Duke & \cellcolor[rgb]{ .906,  .902,  .902}\textcolor[rgb]{ 1,  0,  0}{\textbf{33.9}} & \cellcolor[rgb]{ .906,  .902,  .902}\textcolor[rgb]{ 1,  0,  0}{\textbf{66.7}} \\
\cmidrule{2-9}          & StrongBaseline \cite{kumar2019fairest} & ArXiv'19 & MSMT17 & {43.3}  & {64.5}  & MSMT17 & {36.6}  & {64.8} \\

          & OSNet-IBN \cite{zhou2019omni} & ICCV'19 & MSMT17 & \textcolor[rgb]{ .357,  .608,  .835}{45.6}  & \textcolor[rgb]{ .357,  .608,  .835}{67.4}  & MSMT17  & \textcolor[rgb]{ .357,  .608,  .835}{37.2}  & \textcolor[rgb]{ .357,  .608,  .835}{66.5} \\

          & \cellcolor[rgb]{ .906,  .902,  .902}Baseline & \cellcolor[rgb]{ .906,  .902,  .902}This work & \cellcolor[rgb]{ .906,  .902,  .902}MSMT17 & \cellcolor[rgb]{ .906,  .902,  .902}{39.1} & \cellcolor[rgb]{ .906,  .902,  .902}{60.4} & \cellcolor[rgb]{ .906,  .902,  .902}MSMT17 & \cellcolor[rgb]{ .906,  .902,  .902}{33.8} & \cellcolor[rgb]{ .906,  .902,  .902}{59.9} \\
          & \cellcolor[rgb]{ .906,  .902,  .902}\textbf{SNR(Ours)} & \cellcolor[rgb]{ .906,  .902,  .902}This work & \cellcolor[rgb]{ .906,  .902,  .902}MSMT17 & \cellcolor[rgb]{ .906,  .902,  .902}\textcolor[rgb]{ 1,  0,  0}{\textbf{50.0}} & \cellcolor[rgb]{ .906,  .902,  .902}\textcolor[rgb]{ 1,  0,  0}{\textbf{69.2}} & \cellcolor[rgb]{ .906,  .902,  .902}MSMT17 & \cellcolor[rgb]{ .906,  .902,  .902}\textcolor[rgb]{ 1,  0,  0}{\textbf{41.4}} & \cellcolor[rgb]{ .906,  .902,  .902}\textcolor[rgb]{ 1,  0,  0}{\textbf{70.1}} \\
    \midrule
    \midrule
    \multicolumn{1}{c|}{\multirow{19}[4]{*}{\begin{tabular}[c]{@{}c@{}}Unsupervised\\ Domain\\Adaptation \\(using unlabeled \\ target data)\end{tabular}}} & PTGAN \cite{wei2018person} & CVPR'18 & Market1501 + Duke (U) & --     & 27.4  & Duke + Market1501 (U) & --     & 38.6 \\
          & PUL \cite{fan2018unsupervised} & TOMM'18 & Market1501 + Duke (U) & 16.4  & 30.0    & Duke + Market1501 (U) & 20.5  & 45.5 \\
          & MMFA \cite{lin2018multi} & BMVC'18 & Market1501 + Duke (U) & 24.7  & 45.3  & Duke + Market1501 (U) & 27.4  & 56.7 \\
          & SPGAN \cite{deng2018image} & CVPR'18 & Market1501 + Duke (U) & 26.2  & 46.4  & Duke + Market1501 (U) & 26.7  & 57.7 \\
          & TJ-AIDL \cite{wang2018transferable} & CVPR'18 & Market1501 + Duke (U) & 23.0    & 44.3  & Duke + Market1501 (U) & 26.5  & 58.2 \\
          & ATNet \cite{liu2019adaptive} & CVPR'19 & Market1501 + Duke (U) & 24.9  & 45.1  & Duke + Market1501 (U) & 25.6  & 55.7 \\
          & CamStyle \cite{zhong2018camstyle} & TIP'19 & Market1501 + Duke (U) & 25.1  & 48.4  & Duke + Market1501 (U) & 27.4  & 58.8 \\
          & HHL \cite{zhong2018generalizing} & ECCV'18 & Market1501 + Duke (U) & 27.2  & 46.9  & Duke + Market1501 (U) & 31.4  & 62.2 \\
          & ARN \cite{li2018adaptation} & CVPRW'19 & Market1501 + Duke (U) & 33.4  & 60.2  & Duke + Market1501 (U) & 39.4  & 70.3 \\
          & ECN \cite{zhong2019invariance} & CVPR'19 & Market1501 + Duke (U) & 40.4  & 63.3  & Duke + Market1501 (U) & 43.0    & 75.1 \\
          & UDAP \cite{song2018unsupervised} & ArXiv'18 & Market1501 + Duke (U) & 49.0    & 68.4  & Duke + Market1501 (U) & 53.7  & 75.8 \\
          & PAST \cite{zhang2019self} & ICCV'19 & Market1501 + Duke (U) & \textcolor[rgb]{ .357,  .608,  .835}{54.3} & 72.4  & Duke + Market1501 (U) & 54.6  & 78.4 \\
          & SSG \cite{Fu2018SelfsimilarityGA} & ICCV'19 & Market1501 + Duke (U) & 53.4  & \textcolor[rgb]{ .357,  .608,  .835}{73.0} & Duke + Market1501 (U) & \textcolor[rgb]{ .357,  .608,  .835}{58.3} & \textcolor[rgb]{ .357,  .608,  .835}{80.0} \\
          & \cellcolor[rgb]{ .906,  .902,  .902}Baseline+MAR \cite{yu2019unsupervised} & \cellcolor[rgb]{ .906,  .902,  .902}This work & \cellcolor[rgb]{ .906,  .902,  .902}Market1501 + Duke (U) & \cellcolor[rgb]{ .906,  .902,  .902}{35.2} & \cellcolor[rgb]{ .906,  .902,  .902}{56.5} & \cellcolor[rgb]{ .906,  .902,  .902}Duke + Market1501 (U) & \cellcolor[rgb]{ .906,  .902,  .902}{37.2} & \cellcolor[rgb]{ .906,  .902,  .902}{62.4} \\
          & \cellcolor[rgb]{ .906,  .902,  .902}\textbf{SNR(Ours)}+MAR \cite{yu2019unsupervised} & \cellcolor[rgb]{ .906,  .902,  .902}This work & \cellcolor[rgb]{ .906,  .902,  .902}Market1501 + Duke (U) & \cellcolor[rgb]{ .906,  .902,  .902}\textcolor[rgb]{ 1,  0,  0}{\textbf{58.1}} & \cellcolor[rgb]{ .906,  .902,  .902}\textcolor[rgb]{ 1,  0,  0}{\textbf{76.3}} & \cellcolor[rgb]{ .906,  .902,  .902}Duke + Market1501 (U) & \cellcolor[rgb]{ .906,  .902,  .902}\textcolor[rgb]{ 1,  0,  0}{\textbf{61.7}} & \cellcolor[rgb]{ .906,  .902,  .902}\textcolor[rgb]{ 1,  0,  0}{\textbf{82.8}} \\
\cmidrule{2-9}          & MAR \cite{yu2019unsupervised} & CVPR'19 & MSMT17 + Duke (U) & 48.0    & 67.1  & MSMT17 + Market1501 (U) & 40.0    & 67.7 \\
          & PAUL \cite{yang2019patch} & CVPR'19 & MSMT17 + Duke (U) & \textcolor[rgb]{ .357,  .608,  .835}{53.2}  & \textcolor[rgb]{ .357,  .608,  .835}{72.0}    & MSMT17 + Market1501 (U) & \textcolor[rgb]{ .357,  .608,  .835}{40.1}  & \textcolor[rgb]{ .357,  .608,  .835}{68.5}  \\
          & \cellcolor[rgb]{ .906,  .902,  .902}Baseline+MAR \cite{yu2019unsupervised} & \cellcolor[rgb]{ .906,  .902,  .902}This work & \cellcolor[rgb]{ .906,  .902,  .902}MSMT17 + Duke (U) & \cellcolor[rgb]{ .906,  .902,  .902}{46.2} & \cellcolor[rgb]{ .906,  .902,  .902}{66.3} & \cellcolor[rgb]{ .906,  .902,  .902}MSMT17 + Market1501 (U) & \cellcolor[rgb]{ .906,  .902,  .902}{39.4} & \cellcolor[rgb]{ .906,  .902,  .902}{66.9} \\
          & \cellcolor[rgb]{ .906,  .902,  .902}\textbf{SNR(Ours)} + MAR \cite{yu2019unsupervised} & \cellcolor[rgb]{ .906,  .902,  .902}This work & \cellcolor[rgb]{ .906,  .902,  .902}MSMT17 + Duke (U) & \cellcolor[rgb]{ .906,  .902,  .902}\textcolor[rgb]{ 1,  0,  0}{\textbf{61.6}} & \cellcolor[rgb]{ .906,  .902,  .902}\textcolor[rgb]{ 1,  0,  0}{\textbf{78.2}} & \cellcolor[rgb]{ .906,  .902,  .902}MSMT17 + Market1501 (U) & \cellcolor[rgb]{ .906,  .902,  .902}\textcolor[rgb]{ 1,  0,  0}{\textbf{65.9}} & \cellcolor[rgb]{ .906,  .902,  .902}\textcolor[rgb]{ 1,  0,  0}{\textbf{85.5}} \\
    \bottomrule
    \bottomrule
    \end{tabular}%
  \label{tab:STO}%
\end{table*}%

\begin{table*}[t]
  \centering
  \scriptsize
  \caption{Performance (\%) comparison with the latest domain generalizable ReID method Domain-Invariant Mapping Network (DIMN) \cite{song2019generalizable} under the same experimental setting (\ieno, training on the same five datasets, Market1501\cite{zheng2015scalable}+DukeMTMC-reID\cite{zheng2017unlabeled}+CUHK02\cite{li2013locally}+CUHK03\cite{li2014deepreid}+CUHK-SYSU\cite{xiao2016end}).}
    \begin{tabular}{c|l|cccccccc}
    \toprule
    \multirow{2}[2]{*}{Source} & \multicolumn{1}{c|}{\multirow{2}[2]{*}{Method}} & \multicolumn{2}{c}{Target: PRID} & \multicolumn{2}{c}{Target: GRID} & \multicolumn{2}{c}{Target: VIPeR} & \multicolumn{2}{c}{Target: iLIDs} \\
          &       & mAP   & Rank-1 & mAP   & Rank-1 & mAP   & Rank-1 & mAP   & Rank-1 \\
    \midrule
    \multirow{3}[2]{*}{Market + Duke + CUHK02 + CUHK03 + CUHK-SYSU} & DIMN \cite{song2019generalizable} CVPR'19 & \textcolor[rgb]{ .357,  .608,  .835}{51.9} & \textcolor[rgb]{ .357,  .608,  .835}{39.2} & \textcolor[rgb]{ .357,  .608,  .835}{41.1} & \textcolor[rgb]{ .357,  .608,  .835}{29.3} & \textcolor[rgb]{ .357,  .608,  .835}{60.1} & \textcolor[rgb]{ .357,  .608,  .835}{51.2} & \textcolor[rgb]{ .357,  .608,  .835}{78.4} & \textcolor[rgb]{ .357,  .608,  .835}{70.2} \\
          & \cellcolor[rgb]{ .906,  .902,  .902}Baseline & \cellcolor[rgb]{ .906,  .902,  .902}43.8 & \cellcolor[rgb]{ .906,  .902,  .902}35.0 & \cellcolor[rgb]{ .906,  .902,  .902}37.7 & \cellcolor[rgb]{ .906,  .902,  .902}28.0 & \cellcolor[rgb]{ .906,  .902,  .902}54.6 & \cellcolor[rgb]{ .906,  .902,  .902}45.6 & \cellcolor[rgb]{ .906,  .902,  .902}75.3 & \cellcolor[rgb]{ .906,  .902,  .902}65.0 \\
          & \cellcolor[rgb]{ .906,  .902,  .902}\textbf{SNR (Ours)} & \cellcolor[rgb]{ .906,  .902,  .902}\textcolor[rgb]{ 1,  0,  0}{\textbf{66.5}} & \cellcolor[rgb]{ .906,  .902,  .902}\textcolor[rgb]{ 1,  0,  0}{\textbf{52.1}} & \cellcolor[rgb]{ .906,  .902,  .902}\textcolor[rgb]{ 1,  0,  0}{\textbf{47.7}} & \cellcolor[rgb]{ .906,  .902,  .902}\textcolor[rgb]{ 1,  0,  0}{\textbf{40.2}} & \cellcolor[rgb]{ .906,  .902,  .902}\textcolor[rgb]{ 1,  0,  0}{\textbf{61.3}} & \cellcolor[rgb]{ .906,  .902,  .902}\textcolor[rgb]{ 1,  0,  0}{\textbf{52.9}} & \cellcolor[rgb]{ .906,  .902,  .902}\textcolor[rgb]{ 1,  0,  0}{\textbf{89.9}} & \cellcolor[rgb]{ .906,  .902,  .902}\textcolor[rgb]{ 1,  0,  0}{\textbf{84.1}} \\
    \bottomrule
    \end{tabular}%
  \label{tab:DIMN}%
\end{table*}%

\begin{figure}[t]
  \centerline{\includegraphics[width=0.96\linewidth]{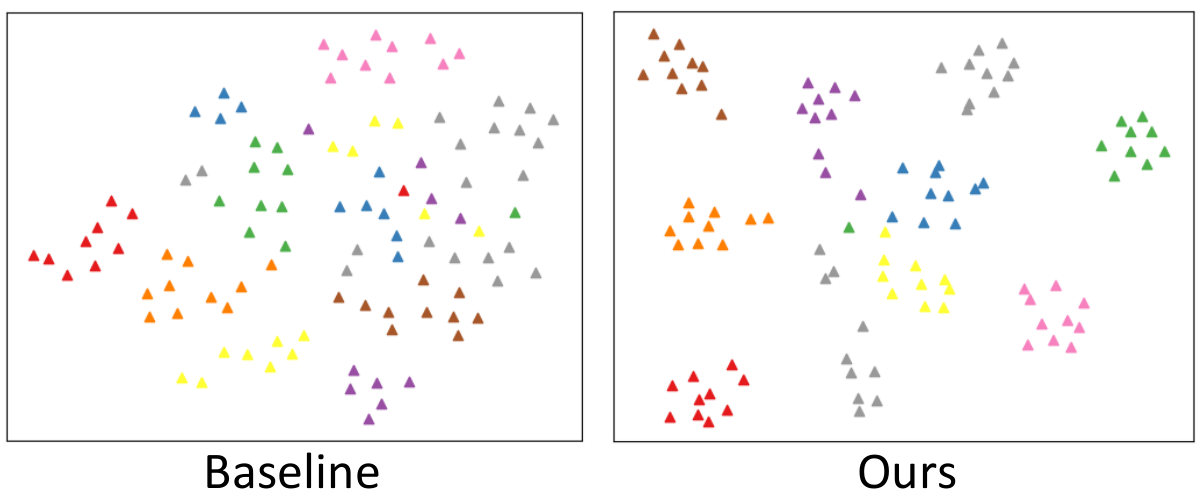}}
  \caption{Visualization of the final ReID feature vector distribution for \emph{Baseline} and \emph{Ours} on the unseen target dataset Duke. Different identities are denoted by different colors.}
\label{fig:vis_tSNE_supp}
\vspace{-8mm}
\end{figure}

\begin{table}
  \centering
  \scriptsize
  \caption{Differences between settings of supervised, domain adaptive, and domain generalizable ReID.}
    \begin{tabular}{ccc}
    \toprule
    Setting & \tabincell{c}{Use target \\domain data?} & \tabincell{c}{Use target \\ domain label?} \\
    \midrule
    Supervised  &  \cmark     &  \cmark \\
    Domain adaptation & \cmark      & \xmark \\
    Domain generalization &  \xmark     &  \xmark \\
    \bottomrule
    \end{tabular}%
  \label{tab:diff}%
\end{table}%

\noindent\textbf{Visualization of ReID Feature Vector Distributions.} In Figure \ref{fig:vis_tSNE_supp}, we further visualize the distribution of the final ReID feature vectors using t-SNE \cite{maaten2008visualizing} for \emph{Baseline} scheme and our final scheme on the unseen target dataset Duke (\ie, Market1501$\rightarrow$Duke). In comparison with \emph{Baseline}, the feature distribution of the same identity (same color) becomes more compact while those of the different identities are pushed away in our scheme. It is easier to distinguish between different identities by our method.

\begin{table*}[t]
  \centering
  \scriptsize
  \caption{Performance (\%) comparisons with the state-of-the-art RGB-IR ReID approaches on SYSU-MM01 dataset. R1, R10, R20 denote Rank-1, Rank-10 and Rank-20 accuracy, respectively.}
    \begin{tabular}{cccccccccccccccccc}
    \toprule
    \multirow{3}[6]{*}{Method} & \multirow{3}[6]{*}{Veneue} & \multicolumn{8}{c}{All Search}                                & \multicolumn{8}{c}{Indoor-Search} \\
\cmidrule{3-18}          &       & \multicolumn{4}{c}{Single-Shot} & \multicolumn{4}{c}{Multi-shot} & \multicolumn{4}{c}{Single-Shot} & \multicolumn{4}{c}{Multi-Shot} \\
\cmidrule{3-18}          &       & mAP   & R1    & R10   & R20   & mAP   & R1    & R10   & R20   & mAP   & R1    & R10   & R20   & mAP   & R1    & R10   & R20 \\
    \midrule
    HOG \cite{dalal2005histograms} & CVPR'05 & 4.24  & 2.76  & 18.3  & 32.0  & 2.16  & 3.82  & 22.8  & 37.7  & 7.25  & 3.22  & 24.7  & 44.6  & 3.51  & 4.75  & 29.1  & 49.4 \\
    MLBP \cite{liao2015efficient} & ICCV'15 & 3.86  & 2.12  & 16.2  & 28.3  & --     & --     & --     & --     & --     & --     & --     & --     & --     & --     & --     & -- \\
    LOMO \cite{liao2015person} & CVPR'15 & 4.53  & 3.64  & 23.2  & 37.3  & 2.28  & 4.70  & 28.3  & 43.1  & 10.2  & 5.75  & 34.4  & 54.9  & 5.64  & 7.36  & 40.4  & 60.4 \\
    GSM \cite{lin2016cross} & TPAMI'17 & 8.00  & 5.29  & 33.7  & 53.0  & --     & --     & --     & --     & --     & --     & --     & --     & --     & --     & --     & -- \\
    One-stream \cite{wu2017rgb} & ICCV'17 & 13.7  & 12.1  & 49.7  & 66.8  & 8.59  & 16.3  & 58.2  & 75.1  & 56.0  & 17.0  & 63.6  & 82.1  & 15.1  & 22.7  & 71.8  & 87.9 \\
    Two-stream \cite{wu2017rgb} & ICCV'17 & 12.9  & 11.7  & 48.0  & 65.5  & 8.03  & 16.4  & 58.4  & 74.5  & 21.5  & 15.6  & 61.2  & 81.1  & 14.0  & 22.5  & 72.3  & 88.7 \\
    Zero-Padding \cite{wu2017rgb} & ICCV'17 & 16.0  & 14.8  & 52.2  & 71.4  & 10.9  & 19.2  & 61.4  & 78.5  & 27.0  & 20.6  & 68.4  & 85.8  & 18.7  & 24.5  & 75.9  & 91.4 \\
    TONE \cite{ye2018hierarchical} & AAAI'18 & 14.4  & 12.5  & 50.7  & 68.6  & --     & --     & --     & --     & --     & --     & --     & --     & --     & --     & --     & -- \\
    HCML \cite{ye2018hierarchical} & AAAI'18 & 16.2  & 14.3  & 53.2  & 69.2  & --     & --     & --     & --     & --     & --     & --     & --     & --     & --     & --     & -- \\
    BCTR \cite{ye2018visible} & IJCAI'18 & 19.2  & 16.2  & 54.9  & 71.5  & --     & --     & --     & --     & --     & --     & --     & --     & --     & --     & --     & -- \\
    BDTR \cite{ye2018visible} & IJCAI'18 & 19.7  & 17.1  & 55.5  & 72.0  & --     & --     & --     & --     & --     & --     & --     & --     & --     & --     & --     & -- \\
    D-HSME \cite{hao2019hsme} & AAAI'19 & 23.2  & 20.7  & 62.8  & 78.0  & --     & --     & --     & --     & --     & --     & --     & --     & --     & --     & --     & -- \\
    cmGAN \cite{dai2018cross} & IJCAI'18 & 27.8  & 27.0  & 67.5  & 80.6  & \textcolor[rgb]{ .357,  .608,  .835}{22.3} & 31.5  & 72.7  & 85.0  & \textcolor[rgb]{ .357,  .608,  .835}{42.2} & \textcolor[rgb]{ .357,  .608,  .835}{31.7} & \textcolor[rgb]{ .357,  .608,  .835}{77.2} & \textcolor[rgb]{ .357,  .608,  .835}{89.2} & \textcolor[rgb]{ .357,  .608,  .835}{32.8} & 37.0  & 80.9  & 92.3 \\
     D$^{2}$RL \cite{wang2019learning} & CVPR'19 & \textcolor[rgb]{ .357,  .608,  .835}{29.2} & \textcolor[rgb]{ .357,  .608,  .835}{28.9} & \textcolor[rgb]{ .357,  .608,  .835}{70.6} & \textcolor[rgb]{ .357,  .608,  .835}{82.4} & --     & --     & --     & --     & --     & --     & --     & --     & --     & --     & --     & -- \\
    \midrule
    Baseline & This work & 25.5  & 26.3  & 66.7  & 80.2  & 19.2  & \textcolor[rgb]{ .357,  .608,  .835}{32.7} & \textcolor[rgb]{ .357,  .608,  .835}{73.5} & \textcolor[rgb]{ .357,  .608,  .835}{86.8} & 39.4  & 30.8  & 75.1  & 86.8  & 29.0  & \textcolor[rgb]{ .357,  .608,  .835}{40.1} & \textcolor[rgb]{ .357,  .608,  .835}{83.1} & \textcolor[rgb]{ .357,  .608,  .835}{93.6} \\
    \textbf{Ours} & This work & \textcolor[rgb]{ 1,  0,  0}{\textbf{33.9}} & \textcolor[rgb]{ 1,  0,  0}{\textbf{34.6}} & \textcolor[rgb]{ 1,  0,  0}{\textbf{75.9}} & \textcolor[rgb]{ 1,  0,  0}{\textbf{86.6}} & \textcolor[rgb]{ 1,  0,  0}{\textbf{27.4}} & \textcolor[rgb]{ 1,  0,  0}{\textbf{41.7}} & \textcolor[rgb]{ 1,  0,  0}{\textbf{83.3}} & \textcolor[rgb]{ 1,  0,  0}{\textbf{92.3}} & \textcolor[rgb]{ 1,  0,  0}{\textbf{50.4}} & \textcolor[rgb]{ 1,  0,  0}{\textbf{40.9}} & \textcolor[rgb]{ 1,  0,  0}{\textbf{83.8}} & \textcolor[rgb]{ 1,  0,  0}{\textbf{91.8}} & \textcolor[rgb]{ 1,  0,  0}{\textbf{40.5}} & \textcolor[rgb]{ 1,  0,  0}{\textbf{50.0}} & \textcolor[rgb]{ 1,  0,  0}{\textbf{91.4}} & \textcolor[rgb]{ 1,  0,  0}{\textbf{96.1}} \\
    \bottomrule
    \end{tabular}%
  \label{tab:RGB-IR-reid}%
\end{table*}%

\section{Comparison with State-of-the-Arts (Complete version)}

To save space, we only present the latest approaches in the paper and here we show comparisons with more approaches in Table \ref{tab:STO}. Besides the description in \emph{Introduction} and \emph{Related Work} sections of our paper, we illustrate the difference between domain generalization and domain adaptation for person ReID in Table \ref{tab:diff}.

Moreover, in Table \ref{tab:DIMN}, we further compare our \emph{SNR} with the latest generalizable ReID method Domain-Invariant Mapping Network (\emph{DIMN}) \cite{song2019generalizable} under the same experimental setting, \ieno, training on the same five datasets, Market1501~\cite{zheng2015scalable} + DukeMTMC-reID~\cite{zheng2017unlabeled} + CUHK02~\cite{li2013locally} + CUHK03~\cite{li2014deepreid} + CUHK-SYSU~\cite{xiao2016end}. We observe that \emph{SNR} not only outperforms the \emph{Baseline} by a large margin (up to 22.7\% in mAP on PRID), but also significantly outperforms \emph{DIMN}\cite{song2019generalizable} by \textbf{14.6\%/6.6\%/1.2\%/11.5\%} in mAP on  PRID/GRID/VIPeR/i-LIDS, respectively.

\section{Performance on Another Backbone}
Our SNR is a plug-and-play module which can be added to available ReID networks. We integrate it into the recently proposed lightweight ReID network OSNet \cite{zhou2019omni} and Table \ref{tab:osnet} shows the results. We can see that by simply inserting SNR modules between the  OS-Blocks, the new scheme \emph{OSNet-SNR} outperforms their best model \emph{OSNet-IBN} by 5.0\% and 5.5\% in mAP for M$\rightarrow$D and D$\rightarrow$M, respectively. Note that, for fair comparison, we use the official released weights and codes \footnote{https://github.com/KaiyangZhou/deep-person-reid} of OSNet \cite{zhou2019omni} to conduct these  experiments.


\section{RGB-Infrared Cross-Modality Person ReID}
To further demonstrate the generalization capability of the proposed SNR module, we conduct experiment on a more challenging RGB-Infrared cross-modality person ReID task, where there is a large style discrepancy between RGB images and Infrared images.

We evaluate our models on the standard benchmark dataset SYSU-MM01 \cite{wu2017rgb}. Following \cite{wu2017rgb}, we conduct evaluation using the released official code based on the average of 10 repeated random split of gallery and probe sets. As shown in Table \ref{tab:RGB-IR-reid}, in comparison with \emph{Baseline}, our scheme which integrates the proposed SNR module on \emph{Baseline} achieves a significant gain of \textbf{8.4\%, 8.2\%, 11.0\%}, and \textbf{11.5\%} in terms of mAP under 4 different experimental settings, and achieves the state-of-the-art performance.

\begin{table}[t]
  \centering
  \footnotesize
  \caption{Evaluation of the generalization capability of proposed SNR modules on OSNet \cite{zhou2019omni}. We use the official released weights and codes of OSNet for the experiments.}
    \begin{tabular}{lcccc}
    \toprule
    \multicolumn{1}{c}{\multirow{2}[2]{*}{Method}} & \multicolumn{2}{c}{M$\longrightarrow$D} & \multicolumn{2}{c}{D$\longrightarrow$M} \\
          & mAP   & Rank-1 & mAP   & Rank-1 \\
    \midrule
    Baseline (ResNet50) & 19.8  & 35.3  & 21.8  & 48.3 \\
    OSNet \cite{zhou2019omni} & 19.3  & 35.2  & 21.7  & 49.9 \\
    OSNet-IBN \cite{zhou2019omni} & \textcolor[rgb]{ .357,  .608,  .835}{26.7}  & \textcolor[rgb]{ .357,  .608,  .835}{48.5}  & \textcolor[rgb]{ .357,  .608,  .835}{26.1}  & \textcolor[rgb]{ .357,  .608,  .835}{57.7} \\
    \textbf{OSNet-SNR} & \textcolor[rgb]{ 1,  0,  0}{\textbf{31.7}} & \textcolor[rgb]{ 1,  0,  0}{\textbf{53.6}} & \textcolor[rgb]{ 1,  0,  0}{\textbf{31.6}} & \textcolor[rgb]{ 1,  0,  0}{\textbf{62.7}} \\
    \bottomrule
    \end{tabular}%
  \label{tab:osnet}%
\end{table}%

\end{document}